\documentclass[11pt]{article}

\usepackage[margin=1in]{geometry}
\usepackage[utf8]{inputenc}
\usepackage[T1]{fontenc}
\usepackage{url}
\usepackage{booktabs}
\usepackage{amsfonts}
\usepackage{nicefrac}
\usepackage{microtype}
\usepackage{xcolor}
\usepackage[colorlinks,linkcolor=magenta,citecolor=blue,urlcolor=blue]{hyperref}

\usepackage{graphicx}
\usepackage{subfigure}
\usepackage{algorithmic}
\usepackage{caption}
\usepackage{amsmath, wasysym, verbatim}
\usepackage{newunicodechar}

\usepackage[linesnumbered,ruled]{algorithm2e}
\DontPrintSemicolon

\usepackage{amsthm}
\usepackage{amssymb,bbm}
\usepackage{mathrsfs}

\usepackage{chngpage}
\usepackage{tabularx}
\usepackage{enumitem}
\usepackage{bm}
\usepackage{mathtools}
\usepackage{cleveref}
\usepackage{eucal}
\usepackage{scalerel}
\usepackage[framemethod=TikZ]{mdframed}

\newcommand*\R[0]{\mathbb{R}}

\newcommand*\pt[0]{\tilde{p}}

\renewcommand*{\Re}{\mathbb{R}}

\newcommand{\real}{\ensuremath{\mathbb{R}}}

\newcommand{\matsnorm}[2]{|\!|\!| #1 | \! | \!|_{{#2}}}
\newcommand{\vecnorm}[2]{\| #1\|_{#2}}

\newcommand{\opnorm}[1]{\ensuremath{\matsnorm{#1}{\tiny{\mbox{op}}}}}

\newcommand{\Exs}{\ensuremath{{\mathbb{E}}}}

\theoremstyle{plain}

\newlength{\widebarargwidth}
\newlength{\widebarargheight}
\newlength{\widebarargdepth}
\DeclareRobustCommand{\widebar}[1]{%
  \settowidth{\widebarargwidth}{\ensuremath{#1}}%
  \settoheight{\widebarargheight}{\ensuremath{#1}}%
  \settodepth{\widebarargdepth}{\ensuremath{#1}}%
  \addtolength{\widebarargwidth}{-0.3\widebarargheight}%
  \addtolength{\widebarargwidth}{-0.3\widebarargdepth}%
  \makebox[0pt][l]{\hspace{0.3\widebarargheight}%
    \hspace{0.3\widebarargdepth}%
    \addtolength{\widebarargheight}{0.3ex}%
    \rule[\widebarargheight]{0.95\widebarargwidth}{0.1ex}}%
  {#1}}

\makeatletter
\long\def\@makecaption#1#2{
        \vskip 0.8ex
        \setbox\@tempboxa\hbox{\small {\bf #1:} #2}
        \parindent 1.5em  
        \dimen0=\hsize
        \advance\dimen0 by -3em
        \ifdim \wd\@tempboxa >\dimen0
                \hbox to \hsize{
                        \parindent 0em
                        \hfil
                        \parbox{\dimen0}{\def\baselinestretch{0.96}\small
                                {\bf #1.} #2
                                }
                        \hfil}
        \else \hbox to \hsize{\hfil \box\@tempboxa \hfil}
        \fi
        }
\makeatother

\long\def\comment#1{}
\definecolor{battleshipgrey}{rgb}{0.52, 0.52, 0.51}
\definecolor{darkgray}{rgb}{0.66, 0.66, 0.66}
\definecolor{darkgreen}{rgb}{0.0, 0.2, 0.13}
\definecolor{darkspringgreen}{rgb}{0.09, 0.45, 0.27}
\definecolor{dukeblue}{rgb}{0.0, 0.0, 0.61}
\definecolor{olivedrab7}{rgb}{0.24, 0.2, 0.12}
\definecolor{darkblue}{rgb}{0.0, 0.0, 0.55}
\definecolor{darkscarlet}{rgb}{0.34, 0.01, 0.1}
\definecolor{candyapplered}{rgb}{1.0, 0.03, 0.0}
\definecolor{ao(english)}{rgb}{0.0, 0.5, 0.0}
\definecolor{applegreen}{rgb}{0.55, 0.71, 0.0}

\theoremstyle{plain}
\newtheorem{theorem}{Theorem}
\newtheorem{lemma}{Lemma}

\newtheorem{proposition}{Proposition}

\newtheorem{remark}{Remark}

\newtheorem{assumption}{Assumption}

\newcommand{\policy}{\pi}

\def\pt{\partial}

\newcommand{\simiid}{\stackrel{\mathrm{i.i.d.}}{\sim}}

\newcommand{\T}{\mathbb{T}}

\newcommand{\ltwospace}{\mathbb{L}^2}

\DeclareFontFamily{U}{mathx}{}
\DeclareFontShape{U}{mathx}{m}{n}{<-> mathx10}{}
\DeclareSymbolFont{mathx}{U}{mathx}{m}{n}
\DeclareMathAccent{\widecheck}{0}{mathx}{"71}

\newenvironment{narrowpara}
  {\par\addvspace{\smallskipamount}\narrower\noindent\ignorespaces}
  {\par\addvspace{\smallskipamount}}

\renewcommand{\tilde}{\widetilde}
\renewcommand{\hat}{\widehat}

\newcommand{\Dset}{\mathcal{D}}

\newcommand{\ffunc}{\ensuremath{f}}
\newcommand{\ctrlE}{H}

\newcommand{\torus}{\mathbb{T}}

\title{Online Imitation Learning for Stabilizing Vlasov--Poisson Plasmas}

\author{%
  Xiaofan Xia\thanks{Department of Statistical Sciences, University of Toronto.}
  \and Qin Li\thanks{Department of Mathematics, University of Wisconsin--Madison.}
  \and Wenlong Mou\thanks{Department of Statistical Sciences, University of Toronto.}
}

\date{}

\begin{document}

\maketitle

\begin{abstract}
  We study the problem of controlling instability in the Vlasov--Poisson plasma systems, a fundamental challenge for nuclear fusion control. Motivated by the partial observability of plasma states in practice, we investigate imitation learning algorithms that learn from a fully-observable expert controller, but operate under macroscopic measurements. We theoretically establish a separation between online and offline imitation learning --- even a small behavior cloning error leads to error compounding that exponentially amplifies instability over time. By way of contrast, the online imitation learning loss can polynomially control the stability of rollout trajectories. We then propose a DAgger-style algorithm that learns the stabilizing controller under partial observability. Simulation results on a 1D Vlasov--Poisson system demonstrate that our algorithm can effectively stabilize plasma instabilities over longer time horizons than behavior cloning.
\end{abstract}

\section{Introduction}
Controlling instability is a central challenge in plasma physics, where small perturbations can lead to rapid energy growth and loss of confinement. In Tokamak fusion reactors, instability can degrade confinement and trigger disruptions, posing a major obstacle to self-sustaining fusion. In practice, feedback control systems---such as those deployed in devices like TCV and DIII-D~\cite{Degrave2022,Seo2024}---operate on macroscopic observables, including density, temperature, and electromagnetic fields. These quantities are accessible to real-time diagnostics and form the basis of most existing control strategies.

However, these observables provide only a low-dimensional summary of the plasma state. At a fundamental level, a plasma is described by a distribution of particles over position and velocity. Many instability mechanisms depend critically on this finer structure, and velocity-space effects have long been recognized as fundamental to plasma dynamics, dating back to the classical work on Landau damping~\cite{Landau1946}. In particular, so-called kinetic instabilities arise from velocity-space features---such as multiple particle populations moving at different speeds---that are not captured by macroscopic moments alone~\cite{Chen2016}.

As a result, stabilization mechanisms for such systems are naturally formulated at the kinetic level, often involving quantities derived from the full distribution function. Recent theoretical work has identified analytically stabilizing control laws in this setting~\cite{EINKEMMER2025113904}. However, these controls rely on information that is not directly observable in practice, creating a fundamental gap between kinetic-level control theory and macroscopic measurement constraints.

The previous work~\cite{xia2026provable} addresses this gap using behavior cloning (BC), which distills the theoretical controller into a history-based policy that operates on macroscopic observations. They also establish theoretical guarantees for the stability of the learned policy, exhibiting adaptivity to the initial condition's complexity. However, their stability upper bound suffers from an exponential dependence on the control time horizon. This motivates the key question:
\begin{quote}
    \textit{When is long-horizon stabilization of Vlasov--Poisson plasmas possible under partial observability, and how can we learn it?}
\end{quote}
In this paper, we exhibit online imitation learning as a key to this question. Our contribution includes
\begin{itemize}
    \item We show that the exponential error compounding of BC is inevitable --- an affine policy with small BC error provably induces an exponential blow-up in the electric energy. This result exhibits a sharp contrast to stabilization of finite-dimensional linear systems, revealing the fundamental instability in phase-mixing systems.
    \item We show that a small online imitation learning loss can polynomially control the stability of rollout trajectories, enabling long-horizon stabilization under partial observability. We also propose a practical DAgger-style algorithm to minimize such loss.
    \item Through numerical simulation, we demonstrate the effectiveness of our algorithm in stabilizing a 1D Vlasov--Poisson system, achieving significantly improved stability over longer time horizons compared to behavior cloning.
\end{itemize}
The rest of this paper is organized as follows. We discuss related work in \Cref{sec:related}.
In \Cref{sec:prelim}, we introduce the Vlasov--Poisson equations and the imitation learning formulation.
In \Cref{sec:theory-alg}, we present the main theoretical results and the proposed algorithm.
In \Cref{sec:simulations}, we report experimental validation. We conclude in \Cref{sec:conclusion} with a discussion of future directions.

\subsection{Related work}\label{sec:related}
This paper connects learning-based plasma control, kinetic control of plasma instabilities, and imitation learning theory.

\paragraph{AI for plasma and fusion control.}
Learning-based methods for tokamak control date back to~\cite{Bishop1995}. Recent years have seen a surge of interest in applying machine learning to plasma control. On the prediction side, data-driven disruption warning has been studied in real-time and multi-machine settings~\cite{Rea2019DIIID,Montes2019Warnings,KatesHarbeck2019}. Related work also uses machine learning to predict specific instability precursors such as locked modes~\cite{Akcay2021LockedMode}. More control-facing systems connect these predictions to actuation: machine-learning feedback has been used on DIII-D for tearing-mode and disruption avoidance~\cite{Fu2020MLControl}, deep disruption prediction has been integrated into the HL-2A plasma control system to trigger mitigation~\cite{Yang2022HL2A}, and control-oriented proximity indicators estimate both disruption probability and remaining time~\cite{Murari2024ControlOriented}. More recently, reinforcement learning has been used to design magnetic shape controllers that transfer from simulators to tokamak reactors~\cite{Degrave2022}, avoid tearing instabilities in high-performance regimes~\cite{Seo2024}, and exploit historical data for offline model-based control~\cite{Char2023}. These works typically learn policies directly from data and treat the plasma as a black-box dynamical system. Our setting is complementary: the expert controller is derived from the kinetic Vlasov--Poisson structure, but the learned policy must act using only partial macroscopic observations.

\paragraph{Control of plasma instabilities.}
Classical tokamak control focuses on magnetic position, current, shape, and profile regulation, using model-based feedback architectures together with equilibrium reconstruction~\cite{AriolaPironti2016,DeTommasi2019}. Much of this pipeline is an inverse problem: plasma states must be inferred from diagnostics, including fast-ion diagnostic tomography~\cite{Prechel2025}, DIII-D equilibrium reconstruction with EFIT, predictive pressure modeling, and Green's-function-augmented machine learning~\cite{Lao2005,Kinsey2022,McClenaghan2024}. Data assimilation extends this inference viewpoint toward adaptive predictive control in fusion experiments~\cite{Morishita2024}. These work typically work with reduced-order fluid models, which are computationally simpler but miss important kinetic effects.

At the kinetic level, the Vlasov--Poisson equation captures velocity-space instability mechanisms that low-dimensional moment descriptions miss~\cite{Landau1946,Penrose1960,Chen2016,MouhotVillani2011}. Mathematical control work has studied exact controllability through interior controls and external force fields~\cite{Glass2003,GlassHanKwan2012}; recent numerical and analytical methods design stabilizing fields through PDE-constrained optimization, particle-in-cell optimal control, and Fourier--Laplace pole elimination~\cite{Einkemmer2024Constrained,Bartsch2024,EINKEMMER2025113904}. These approaches generally assume access to the distribution function or solve an inverse design problem for the external field. We instead start from a kinetic expert and ask whether its stabilizing action can be reliably distilled into a partially observed feedback policy.
The closest prior work is the recent preprint~\cite{xia2026provable}, which studies behavior cloning for this problem and proves guarantees that adapt to the initial condition's complexity. Their stability bound, however, has exponential horizon dependence.

\paragraph{Online versus offline imitation learning.}
The separation between behavior cloning (BC) and interactive imitation learning is a central theme in imitation learning theory. Early discrete-time reductions showed that supervised BC is exposed to covariate shift and can suffer horizon-squared compounding, while online data aggregation and no-regret algorithms such as DAgger train on learner-induced state distributions and give stronger long-horizon guarantees~\cite{RossBagnell2010,RossGordonBagnell2011,RossBagnell2014}. Later work refined this picture: \cite{Rajaraman2020} shows statistical suboptimality of BC in tabular MDPs, while Foster et al.~\cite{Foster2024} show that BC can achieve near-optimal horizon dependence when trained with log loss. For continuous-state discrete-time dynamical systems, Simchowitz et al.~\cite{Simchowitz2025} prove strong negative results for BC when exponential stability is absent.
Our results add a kinetic PDE perspective: in Vlasov--Poisson stabilization, BC suffers exponential error compounding even when the controlled system is exponentially stable, whereas controlling the online imitation loss along the learner's own rollout yields polynomial error compounding.

\section{Preliminaries: Vlasov--Poisson System and Imitation Learning}\label{sec:prelim}

\subsection{Vlasov--Poisson equations}

In fusion applications, plasmas are typically confined in toroidal or
axisymmetric devices (such as tokamaks). These geometries are well approximated
by a circular configuration, for which a one-dimensional periodic model is
standard. Accordingly, we work on $x\in[0,L_{x}]$ equipped with periodic
boundary conditions. The dynamics are governed by the one-dimensional VP
system:

\begin{equation}\label{eq:vlasov-poisoon_system_ext_1d}
\left\{\begin{array}{l}
   \partial_{t}f_t + v\,\partial_{x}f_t - (E_t+H_t)\,\partial_{v}f_t = 0\,,\\
   E_t = \partial_{x}V_t\,, \\
   \partial_{xx} V_t = 1 - \rho_t = 1 - \displaystyle\int f_t\,\mathrm{d}v\,.
\end{array}\right.
\end{equation}
Here $f_t(x,v)$ denotes the particle distribution at time $t$. The dynamics are driven by the total electric field $E_t+H_t$, where $E_t$ is the self-consistent field induced by the charge density
\[
\rho_t(x) = \int f_t(x,v)\,\mathrm{d}v,
\]
through the Poisson equation and the relation $E_t=\partial_x V_t$.
The system is initialized close to an equilibrium distribution $\mu(v)$ with a small spatial perturbation:
\begin{equation}\label{eq:perturbation:example1}
f_0(x,v)
=
\left(1+\varepsilon f_p(x)\right)\mu(v),
\end{equation}
with $\mu(v)$ denoting the equilibrium state that one wishes to maintain and the perturbation is introduced as a physical space variation multiplication at the order of $\varepsilon$. Due to the introduction of this perturbation, the system will develop turbulence, leading to an undesirable exponential growth of electric energy. The external field $H$ is introduced to suppress such growth and maintain stability.

\paragraph{Expert stabilization and phase mixing.}

Consider the expert controller
\(
H_t^*=-E_t^*,
\)
which exactly cancels the self-consistent field. The dynamics reduce to free transport:
\(
f_t^*(x,v)
=
f_0(x-vt,v).
\)
The resulting phase mixing yields decay of the electric field. To quantify this decay, we impose the following regularity condition on the initial distribution.
For $q\ge2$, define
\begin{align*}
\|f\|_{\mathcal C_q^m}
&\coloneqq
\sup_{\bm 1^\top\alpha=m}
\sup_{(x,v)\in\torus\times\mathbb R}
(1+|v|)^q
|\partial^\alpha f(x,v)|,
\qquad
\|f\|_{\mathcal E_q(\lambda)}
\coloneqq
\sum_{m=0}^\infty
\frac{\lambda^m}{m!}
\|f\|_{\mathcal C_q^m}.
\end{align*}
\begin{assumption}[Initial regularity]
\label{assume:initial-regularity}
There exist constants $\lambda_0,B_0>0$ and $q\ge2$ such that
\[
\|f_0\|_{\mathcal E_q(\lambda_0)}
\le
B_0
\]
almost surely.
\end{assumption}
Assumption~\ref{assume:initial-regularity} is a standard analyticity and velocity-decay condition used in the study of Landau damping~\cite{MouhotVillani2011}. It is satisfied, for example, by smooth spatial densities coupled with Gaussian-mixture velocity distributions.

The resulting phase mixing yields exponential electric-field decay.

\begin{proposition}[Phase-mixing {\cite{xia2026provable}}]
\label{prop:expert-decay}
Under Assumption~\ref{assume:initial-regularity}, there exist constants
$c_1,c_2>0$ such that
\[
\|E_t^*\|_{\ltwospace_x}^2
\le
c_1 e^{-c_2 t},
\qquad
t\ge0.
\]
\end{proposition}
\paragraph{Interaction and observation model:} Let $\nu$ be a probability distribution over phase-space densities. We assume that for each run of the system, an initial density $f_0$ is independently drawn from $\nu$. The system then evolves according to the Vlasov--Poisson equations~\eqref{eq:vlasov-poisoon_system_ext_1d} with the control field $H$. At each time $t$, the controller observes a history of macroscopic measurements $\Dset_t[f_s: 0\leq s \leq t]$ that are functionals of the phase-space density trajectory up to time $t$. For example, the previous work~\cite{xia2026provable} considered a collection of density sensors placed at $(x_i)_{i = 1}^N$, the controller may observe
\begin{align*}
    \Dset_t[f_s: 0\leq s \leq t] = \bigg[ \int f_{k \eta} (x_i, v) dv + \xi_{i, k}  \bigg]_{i \in [N], k \eta \in [0, t]}, \quad \mbox{where } \xi_{i, k} \simiid \mathcal{N}(0, \sigma^2).
\end{align*}
Our framework accommodates general observation functionals satisfying mild regularity conditions, which applies to a wide range of practical measurement setups.

\subsection{Imitation learning for plasma control}
Given an expert policy $\policy^*$ that observes the full phase-space density, we seek to learn a policy $\policy\in\Pi$ that acts on macroscopic observation histories. For any trajectory $g_{0:t}=\{g_s:0\le s\le t\}$, we write
\[
H_t^\policy[g]
\coloneqq
\policy(\Dset_t[g_s:0\le s\le t]),
\qquad
H_t^*[g]
\coloneqq
\policy^*(g_t).
\]
We consider offline and online imitation learning algorithms, which aim to minimize the following losses, respectively:
\begin{itemize}
    \item \textbf{Behavior cloning (BC) loss:} 
   \begin{align}
\mathcal{L}_{\mathrm{BC}}(\policy)
=
\int_0^T
\Exs\Big[
\vecnorm{
H_t^\policy[f^*] - H_t^*[f^*]
}{\ltwospace}^2
\Big]dt .
\label{eq:bc-loss}
\end{align}
    \item \textbf{Online imitation loss:}
    \begin{align}
\mathcal{L}_{\mathrm{online}}(\policy)
=
\int_0^T
\Exs\Big[
\vecnorm{
H_t^\policy[f^\policy] - H_t^*[f^\policy]
}{\ltwospace}^2
\Big]dt .
\label{eq:online-loss}
\end{align}
\end{itemize}
Here $f_t^*$ and $f_t^\policy$ denote the phase-space densities at time $t$ under the expert policy and the learned policy, respectively. The BC loss evaluates the learned policy only on expert-induced observation histories, whereas the online imitation loss evaluates it on its own rollout distribution. Given a class $\Pi$ of policies, we seek to find $\policy \in \Pi$ that minimizes these losses.
\section{Theoretical results and algorithms}\label{sec:theory-alg}
This section establishes a sharp separation between offline and online imitation learning for plasma stabilization. We first show that behavior cloning can exhibit exponential instability under closed-loop deployment despite arbitrarily small imitation error. We then prove that online imitation learning controls the rollout instability polynomially and introduce a practical DAgger-style algorithm. We start by contrasting the fundamental limitations of open-loop behavior cloning (BC) with the theoretical guarantees of online imitation learning algorithms such as DAgger. We then introduce a practical DAgger-based algorithm for stabilizing Vlasov--Poisson plasmas.

\subsection{Limitation of behavior cloning}
\label{subsec:bc-limitation}
In this section, we establish a fundamental limitation of open-loop behavior cloning (BC) in the Vlasov--Poisson system. We show that even arbitrarily small training error on the expert-induced distribution can lead to exponential instability under closed-loop deployment.

We first construct a regime in which behavior cloning exhibits exponential instability despite arbitrarily small offline imitation error.
\paragraph{Setup.}
Consider the Vlasov--Poisson system on the periodic torus
$\torus=[0,L]$ with the symmetric two-stream equilibrium
\[
\mu(v)
=
\frac{1}{2\sqrt{2\pi}\sigma}
\left(
e^{-(v-v_0)^2/(2\sigma^2)}
+
e^{-(v+v_0)^2/(2\sigma^2)}
\right),
\]
where $(v_0,\sigma)=(2.0,0.5)$. This equilibrium lies in the classical two-stream instability regime.

Define the Penrose parameter
\(
M
\coloneqq
\int_{\mathbb R}\frac{\mu'(v)}{v}\,dv.
\)
For $(v_0,\sigma)=(2.0,0.5)$, we obtain numerically
\(
M\approx0.3263.
\)
Choosing
\(
L=\frac{5\pi}{\sqrt M},
\)
then admissible Fourier modes $k_n=2\pi n/L$ satisfy
\(
k_1^2<M,
k_2^2<M,
k_3^2>M.
\)
Hence only the first two modes are linearly unstable.

For each spatial mode $k$, the linearized V--P equation admits normal modes $e^{\lambda t}e^{ikx}$, where $\lambda$ satisfies
\begin{align}
\mathcal D_k(\lambda)
=
1-\frac1{k^2}
\int_{\mathbb R}
\frac{\mu'(v)}{v-i\lambda/k}\,dv
=0.
\label{eq:dispersion-complex}
\end{align}
Classical Penrose theory implies that unstable roots exist precisely when
$k^2<M$. For the symmetric two-stream equilibrium above, the unstable root is unique in the open right half-plane and is real. Solving \eqref{eq:dispersion-complex} numerically gives
\(
\gamma_1\approx0.296,
\gamma_2\approx0.257,
\)
so that $\gamma_i$ is the unique root of $\mathcal D_{k_i}$ with positive real part for $i=1,2$.

\paragraph{Expert trajectory.}
Initialize the system from
\[
f_0(x,v)
=
\mu(v)\bigl(1+\zeta\cos(k_1x+\Theta)\bigr),
\]
where $|\zeta|\le\epsilon_0$ almost surely and
$\Theta\sim\mathrm{Unif}[0,2\pi)$.

Under the expert controller $H_t^*=-E_t^*$, the dynamics reduce to free transport. Consequently, the expert trajectory remains supported entirely on the $k_1$ mode, and therefore
\(
E_{|k|=k_2,t}^*
\equiv0.
\)

We now construct a policy that approximates the expert with small BC error. For the lower bound, we allow the policy to depend on the full state, and we construct an affine policy. In particular, we let
\[
H_t^\pi
=
-E_{|k|=k_1,t}^\pi
+
\epsilon_{\mathrm{tr}}\varphi_2(x),
\]
where
\(
\varphi_2(x)
=
\sqrt{2/L}\cos(k_2x)
\)
is the normalized $k_2$ Fourier mode.

Since the expert trajectory never excites the $k_2$ mode, the policy error
along the expert rollout is exactly
\(
\|H_t^\pi-H_t^*\|_{\ltwospace_x}
=
\epsilon_{\mathrm{tr}}.
\)
Hence the behavior cloning loss is arbitrarily small, yet the resulting
closed-loop dynamics become exponentially unstable.roduces a small bias on the unobserved $k_2$ mode:\looseness=-1

\begin{theorem}[Small BC error induces exponential instability]
\label{thm:bc-limitation}
Consider the randomized initialization above and the closed-loop Vlasov--Poisson dynamics driven by the policy $H_t^\pi$. For all sufficiently small $\epsilon_{\mathrm{tr}}>0$, provided
\(
\epsilon_0 \le c_\epsilon \log^{-4}(1/\epsilon_{\mathrm{tr}}),
\)
there exist constants $\delta_0,a_0,a_1>0$ such that:
\begin{itemize}
    \item the behavior cloning loss satisfies
    \(
    \mathcal{L}_{\mathrm{BC}}(\pi)=\epsilon_{\mathrm{tr}}^2;
    \)
    \item Let $T^*
=
\frac1{\gamma_2}
\log\big(
\frac{\delta_0}{a_0\epsilon_{\mathrm{tr}}}
\big)$, for every realization of $(\zeta,\Theta)$, the electric field satisfies
\[
    \|E_t^\pi\|_{\ltwospace_x}
    \ge
    \frac12 a_0 \epsilon_{\mathrm{tr}} e^{\gamma_2 t}
    -
    a_1 \epsilon_{\mathrm{tr}} e^{\gamma_2 t/2},
    \qquad
    t\in[0,T^*].
\]
\end{itemize}
In particular,
\(
\|E_{T^*}^\pi\|_{\ltwospace_x}
=
\Omega(1),
\)
despite the exponentially decaying expert trajectory.
\end{theorem}
See \Cref{app:bc-limitation} for the detailed proof.
\begin{remark}[Generality]
\label{rem:universality}
Theorem~\ref{thm:bc-limitation} extends to any equilibrium admitting an unstable mode absent from the expert trajectory. A small BC error along such a direction induces exponential instability.
\end{remark}
\paragraph{Interpretation.}
Theorem~\ref{thm:bc-limitation} shows that BC fails to control unobserved unstable modes. Although the expert trajectory remains confined to the stable $k_1$ manifold, a vanishing error along the unseen $k_2$ direction is exponentially amplified by the Vlasov--Poisson dynamics, producing severe covariate shift and macroscopic instability.
This phenomenon can be interpreted as a severe form of covariate shift: the deployment trajectory leaves the expert-induced distribution and enters dynamically unstable regions that were never seen during training.
\paragraph{Proof sketch.}
Let $g_t=f_t^\pi-f_t^*$ denote the deviation from the expert trajectory. The perturbation dynamics are driven by the forcing term $\epsilon_{\mathrm{tr}}\varphi_2(x)$. Projecting onto the unstable $k_2$ eigenmode yields
\[
\frac{d}{dt}A_2(t)
=
\gamma_2 A_2(t)+c_2,
\]
so that $A_2(t)\sim e^{\gamma_2 t}$. A bootstrap argument in $\mathbb H^3_{x,v}$ shows that nonlinear terms remain perturbative up to
\(
T^*\sim \log(1/\epsilon_{\mathrm{tr}}),
\)
allowing the unstable linear mode to dominate and drive the electric field to $\Omega(1)$ scale.

Focusing on discrete-time dynamical systems in continuous state-action spaces, a recent work~\cite{Simchowitz2025} establishes a similar exponential blowup for BC and other offline imitation learning algorithms. They also show that such blow-up will not happen if the system satisfies a strong exponential stability condition. The controlled Vlasov--Poisson system exhibits exponential decaying electric energy under the expert policy. However, different from the spectral stability phenomenon in classical linear systems, the Vlasov--Poisson system is a \emph{phase-mixing} system, where the distribution function develops finer and finer velocity-space structures over time. This phase-mixing mechanism creates a fundamental instability that amplifies small errors in the control field, even when the expert trajectory itself is exponentially stable. For comparison, we will study a finite-dimensional linear system, where small BC error guarantees stability.

\paragraph{Comparison to finite-dimensional linear systems} \Cref{thm:bc-limitation} shows that even an arbitrarily small behavior cloning loss could be amplified to a macroscopic blow-up in the Vlasov--Poisson system. This phenomenon is fundamentally different from the finite-dimensional linear system case, where behavior cloning can guarantee stability under sufficient state coverage. To illustrate this contrast, we consider the following controlled linear dynamical system in $\real^d$:
\begin{align*}
    \dot{x}_t = A x_t + B u_t, \qquad x_0 \sim \nu_0,
\end{align*}
where $A \in \real^{d \times d}$, $B \in \real^{d \times m}$, $u_t \in \real^m$, and $\nu_0$ has a finite second moment. The only randomness is the initial condition $x_0$; conditional on $x_0$, the trajectory is generated by the deterministic ODE. Suppose that the target control policy takes the form $u^*(x) = K^* x + b^*$ for some $K^* \in \real^{m \times d}, b^* \in \real^m$. The expert trajectory is the random process induced by
\begin{align*}
    \frac{d}{dt} x_t^* = (A + B K^*) x_t^* + B b^*, \qquad x_0^* = x_0.
\end{align*}
If $A + B K^*$ satisfies the Lyapunov stability criterion, i.e., there exists a positive definite matrix $P_0$ such that $(A + B K^*)^\top P_0 + P_0 (A + B K^*) \preceq - \rho_0 I_d$ for some $\rho_0 > 0$, then the expert trajectory $x_t^*$ is globally exponentially stable, and the system converges to the unique equilibrium $x_\infty^* = -(A + B K^*)^{-1} B b^*$.

Let us consider a class of affine policies $\widehat{u}(x) = \widehat{K} x + \widehat{b}$, where $\widehat{K} \in \mathbb{R}^{m \times d}$ and $\widehat{b} \in \mathbb{R}^m$. The controlled trajectory under such a policy is given by
\begin{align*}
    \frac{d}{dt}{\widehat{x}}_t = (A + B \widehat{K}) \widehat{x}_t + B \widehat{b}, \qquad \widehat{x}_0 = x_0.
\end{align*}
The following proposition shows that small behavior cloning error is sufficient to guarantee exponential stability of the learned policy, without any exponential blow-up in the error.
\begin{proposition}\label{prop:bc-lin-finite}
For any horizon $T > 0$, define
\begin{align*}
    \widebar{x}_T \coloneqq \frac{1}{T} \int_0^T \Exs[x_t^*] dt,
    \qquad
    \Sigma_T \coloneqq \frac{1}{T} \int_0^T \Exs \big[(x_t^* - \widebar{x}_T) (x_t^* - \widebar{x}_T)^\top\big] dt.
\end{align*}
All expectations are over the random initialization $x_0 \sim \nu_0$. Assume that there exists $\lambda_0 > 0$ such that $\Sigma_T \succeq \lambda_0 I_d$.
Then there exist constants $c_1, c_2 > 0$ and $\bar{\rho} > 0$ depending only on $(K^*, b^*, A, B, P_0, \rho_0, \lambda_0)$ and the distribution of $x_0$, but not on $T$, such that for any affine policy $\widehat{u}(x) = \widehat{K} x + \widehat{b}$ satisfying $\mathcal{L}_{\mathrm{BC}}(\widehat{u}) \leq c_1$, the closed-loop trajectory $\widehat{x}_t$ satisfies
\begin{align*}
    \Exs \big[ \vecnorm{\widehat{x}_t - x_\infty^*}{2}^2 \big]
    \leq c_2 e^{-\bar{\rho} t} \Exs \big[ \vecnorm{x_0 - x_\infty^*}{2}^2 \big]
    + c_2 \mathcal{L}_{\mathrm{BC}}(\widehat{u}),
     \qquad \forall t \in [0,T].
\end{align*}
\end{proposition}
\noindent See \Cref{app:bc-lin-finite} for the proof of this proposition.
\subsection{Theoretical guarantees for online imitation learning}
\label{subsec:dagger-guarantees}
The instability in Theorem~\ref{thm:bc-limitation} arises from covariate shift: behavior cloning is trained only on the expert trajectory and therefore does not control errors along unstable rollout directions. Online imitation learning methods such as DAgger instead train on the learner-induced trajectory distribution.

Let
\(
\epsilon_{\mathrm{tr}}^2
\coloneqq
\mathcal{L}_{\mathrm{online}}(\pi)
\)
denote the online imitation loss of a policy $\pi$. We show that controlling this loss reduces error amplification from exponential to polynomial growth over long horizons.

We impose the following regularity condition on the learned controller.

\begin{assumption}[Policy regularity]
\label{assume:policy-class}
The control field $H_t^\pi$ is uniformly Lipschitz in space:
\[
\|\partial_x H_t^\pi\|_{\mathbb L^\infty}
\le
L_1,
\qquad
\forall t\ge0.
\]
\end{assumption}

The following theorem gives the resulting long-horizon stability guarantee.
\begin{theorem}[Polynomial error growth via online imitation]
\label{thm:stability-via-excess-risk}
Let $\pi$ satisfy Assumption~\ref{assume:policy-class}, and let the initial condition $f_0$ satisfy Assumption~\ref{assume:initial-regularity}. Fix a scalar $\eta\in(0,1)$ and define
\(
\epsilon_{\mathrm{tr}}^2
=
\mathcal{L}_{\mathrm{online}}(\pi),
\epsilon
=
\frac{\epsilon_{\mathrm{tr}}}{\sqrt{\eta}},
\). Then there exist constants $C_1,C_2,c_*>0$, depending only on the equilibrium distribution, initial regularity, and policy regularity constant, such that with probability at least $1-\eta$, we have: for all
\(
0\le t\le T^*
\coloneqq
c_*\epsilon^{-2/9},
\)
the electric field generated by $\pi$ satisfies
\[
\|E_t^\pi\|_{\ltwospace_x}
\le
C_1 e^{-C_2 t}
+
C_1 \epsilon^{1/2}(1+t)^{9}.
\]
In particular, the electric field exhibits at most polynomial growth over a polynomial time horizon in $\epsilon_{\mathrm{tr}}^{-1}$ and does not undergo the exponential amplification of Theorem~\ref{thm:bc-limitation}.
\end{theorem}
See Appendix~\ref{app:stability-via-excess-risk} for the proof of this theorem. \Cref{thm:stability-via-excess-risk} shows that controlling the online imitation loss yields a polynomial stability guarantee, in contrast to the exponential instability of behavior cloning. Therefore, optimizing the online imitation loss is crucial for stabilizing the Vlasov--Poisson system under partial observations.

\paragraph{Proof sketch.}
Let
\(
    D(t)
    =
    \big(
    \int_0^t
    \|\delta_s\|_{L^\infty(\mathbb T)}^2\,ds
    \big)^{1/2},
\)
where
$\delta_t=\ctrlE_t^\pi[f^\pi]-\ctrlE_t^*[f^\pi]$ denotes the online
policy error, with both actions evaluated along the learner rollout.
The online imitation loss controls $\delta_t$ only in time-integrated
$L^2_x$, while the Vlasov dynamics require $L^\infty_x$ control. Using the one-dimensional Gagliardo--Nirenberg inequality together with the
Poisson structure, we derive a bootstrap inequality showing that if
$D(t)\le1$, then in fact
\(
    D(t)
    \lesssim
    \epsilon_{\mathrm{tr}}\mathrm{poly}(t).
\)
Choosing
\(
    T^*=\mathrm{poly}(1/\epsilon_{\mathrm{tr}})
\)
sufficiently small closes the bootstrap by continuity. The density estimate
then propagates through the Poisson equation, yielding the claimed polynomial
upper bound for the electric field on $[0,T^*]$.
\subsection{A DAgger-style algorithm}
To minimize the online imitation loss~\eqref{eq:online-loss}, we consider a DAgger-style algorithm that iteratively solves the fixed-point problem. In particular, given a fixed policy $\pi$, we define the population-level one-step loss functional
\[
\mathcal{L}(\pi';\pi)
=
\int_0^T
\Exs\Big[
\|H_t^{\pi'}[f^\pi]-H_t^*[f^\pi]\|_{\ltwospace}^2
\Big]dt,
\]
where the expectation is over the random initialization and the trajectory induced by $\pi$ and $f^\pi$ denotes the trajectory induced by policy $\pi$. The DAgger algorithm iteratively updates the policy by approximately solving the following optimization problem:
\begin{align*}
    \pi_{n+1} = \arg\min_{\pi' \in \Pi} \mathcal{L} (\pi'; \pi_n), \quad \mbox{for } n = 0, 1, 2, \ldots
\end{align*}
which can be solved by standard supervised learning methods. In practice, we use a mixture policy in the rollout to ensure stability of the learning process, and the iterate becomes
\begin{align*}
    \pi_{n+1} = \arg\min_{\pi' \in \Pi} \mathcal{L} (\pi'; (1 - \alpha_n) \pi_n + \alpha_n \pi^*),\quad \mbox{for } n = 0, 1, 2, \ldots
\end{align*}
where $\alpha_n \in (0,1)$ is a mixing parameter.
\section{Experimental Results}\label{sec:simulations}
We empirically study the control performance of offline behavior cloning and online imitation learning
on the one-dimensional Vlasov--Poisson system. In particular, we show that offline behavior cloning induces unstable
long-horizon plasma dynamics, whereas online imitation learning via DAgger
substantially improves long-term stability.

\begin{figure*}[htbp]
    \centering
    \includegraphics[width=0.8\textwidth]{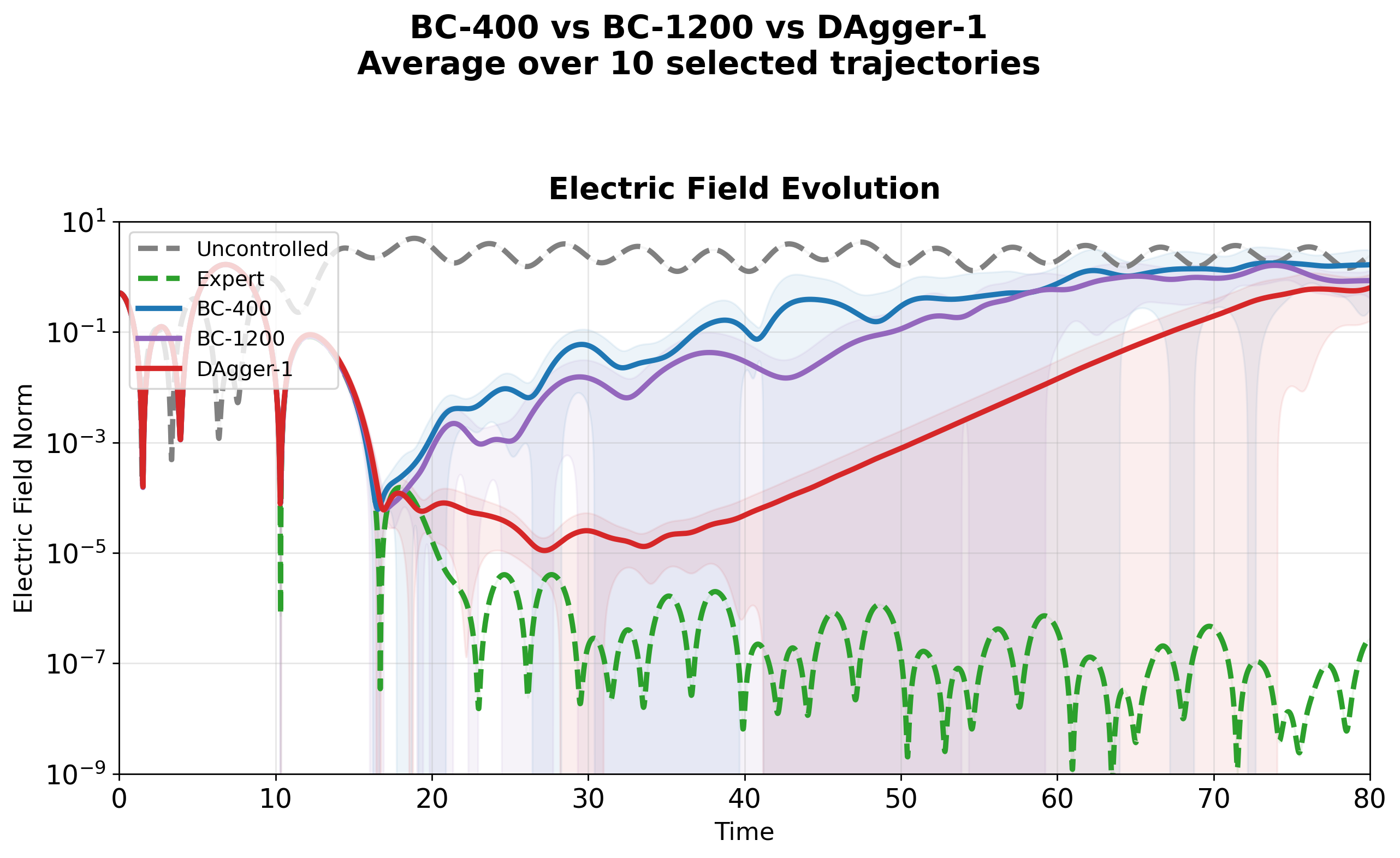}
    \caption{
    Evolution of the electric energy $\mathcal{E}(t) = \frac{1}{2}\|E_t\|_{\mathbb{L}^2_x}^2$ for the
    uncontrolled dynamics, the analytic expert, BC policies trained on $400$
    and $1200$ offline trajectories, and DAgger-1. Curves show pointwise
    empirical means over $10$ rollout trajectories with independently sampled
    random initial phases, and shaded regions denote one pointwise standard
    deviation across those trajectories. Both BC policies eventually exhibit unstable energy growth, while one round
of online aggregation substantially delays the instability and keeps the
rollout closer to the expert trajectory over long horizons.
    }
    \label{fig:field_bc_vs_dagger}
\end{figure*}
\subsection{Experimental Setup}
\paragraph{Dynamics and instability regime.}
We consider the one-dimensional Vlasov--Poisson system on a periodic domain
of length $L=10\pi$ with a symmetric two-stream equilibrium
$(\bar v,\sigma)=(2.4,1.0)$.

All trajectories are initialized from the same dual-mode perturbation:
\[
f_0(x,v)
=
\mu(v)
\Big(
1
+
\varepsilon\cos(k_1x+\phi_1)
+
\varepsilon\cos(k_2x+\phi_2)
\Big),
\]
where $\mu(v)$ is the two-stream equilibrium distribution,
$k_1=1$ and $k_2=5$ are unstable Fourier modes,
$\phi_1,\phi_2$ are random phases,
and $\varepsilon=0.05$ is the perturbation amplitude.
\paragraph{Experimental interface.}
The learner maps a 50-step history from four density sensors, together with
temporal differences, to the first eight Fourier modes of the control field;
it observes neither the full phase-space distribution nor the electric field.
\paragraph{Policies and baselines.}
The main-text results compare:
\begin{itemize}
    \item \textbf{BC-400:} offline behavior cloning trained on $400$ expert trajectories.
    \item \textbf{BC-1200:} offline behavior cloning trained on $1200$ trajectories.
    \item \textbf{DAgger-1:} initialized from BC-400 and fine-tuned using one DAgger iteration with $800$ additional aggregated trajectories.
    \item \textbf{Baselines:} the analytic expert and uncontrolled dynamics.
\end{itemize}

\paragraph{Evaluation metrics.}
We evaluate rollout stability through two physically meaningful diagnostics:
(i) the macroscopic electric energy $\mathcal{E}(t) = \frac{1}{2}\|E_t\|_{\mathbb{L}^2_x}^2$, which directly
measures instability growth, and
(ii) phase-space density snapshots $f_t(x,v)$ visualizing the microscopic
evolution of the plasma.

\paragraph{Statistical reporting.}
Electric-energy curves report empirical means over $10$ rollout trajectories
with independently sampled random phases $(\phi_1,\phi_2)$; shaded regions
denote one standard deviation. Phase-space snapshots show a representative
rollout whose final-time electric-energy value is closest to the median
across all sampled trajectories.


\subsection{Electric-Field Evolution}

Figure~\ref{fig:field_bc_vs_dagger} shows that both BC policies eventually
develop unstable electric-energy growth despite low imitation error on expert
data. Increasing the offline dataset from $400$ to $1200$ trajectories delays the
onset of instability but does not remove it. In contrast, DAgger-1
maintains substantially smaller electric energies over long horizons.
Appendix~\ref{app:expanded-evaluation} provides stronger baseline comparisons
and further evaluates the robustness of our findings across online
aggregation rounds and kinetic-instability regimes.
\subsection{Phase-Space Evolution}
\begin{figure*}[htbp]
    \centering
    \includegraphics[width=\textwidth]{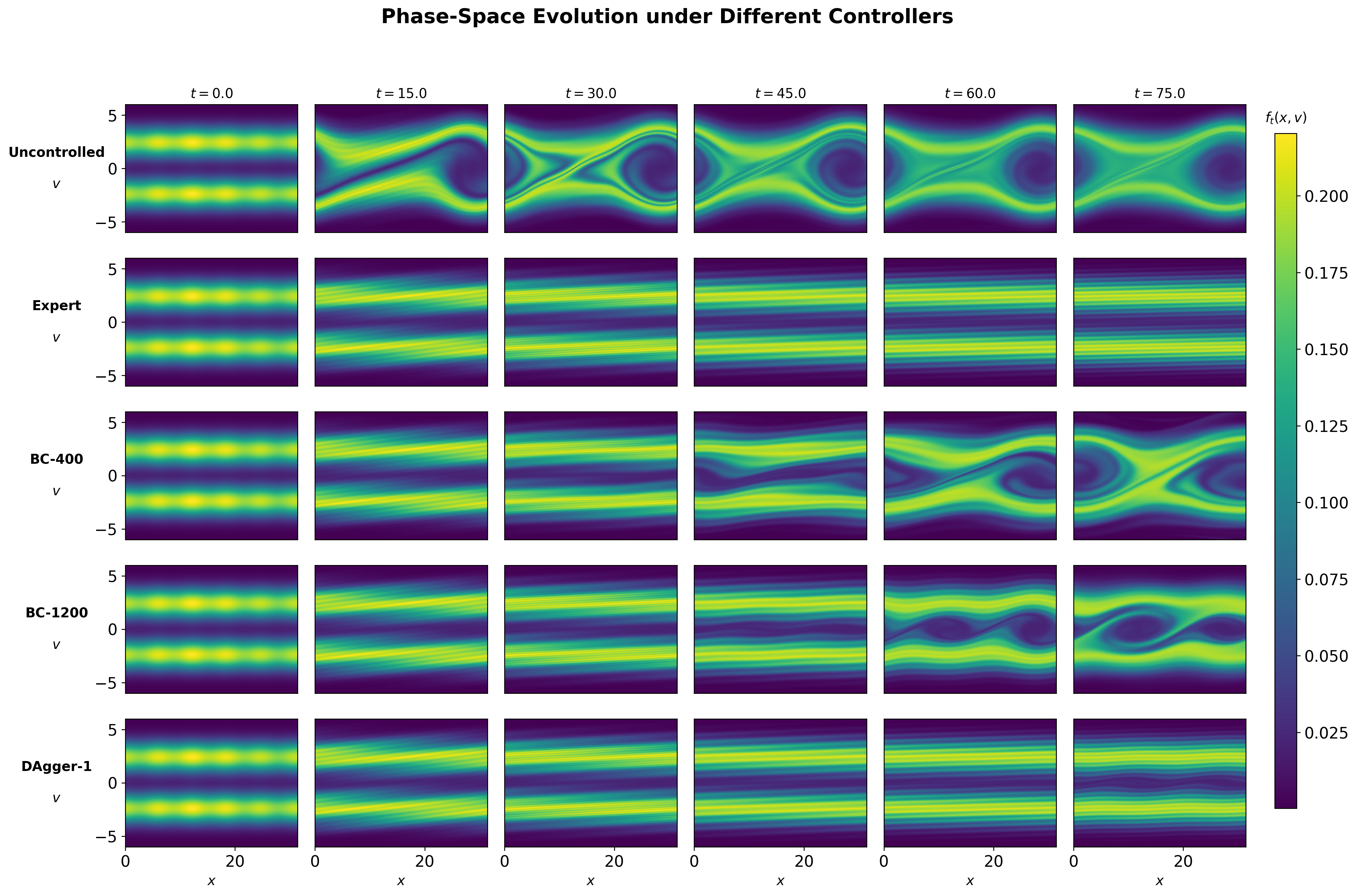}
    \caption{
    Phase-space density snapshots $f_t(x,v)$ at times
    $t \in \{0,15,30,45,60,75\}$ for a representative rollout trajectory
    under five controllers: uncontrolled, expert, BC-400, BC-1200, and
    DAgger-1. The representative case is chosen so that its final-time
    electric-energy value is closest to the median across the sampled
    rollout cases. The uncontrolled dynamics quickly develop large-scale
    deformation and filamentation. BC-400 eventually reproduces this unstable
    behavior, while BC-1200 delays but does not remove the onset of kinetic
    distortion. DAgger-1 remains visibly closer to the expert evolution
    throughout the rollout, with substantially less phase-space twisting and
    mode excitation at late times.
    }
    \label{fig:heatmap_bc_vs_dagger}
\end{figure*}
Figure~\ref{fig:heatmap_bc_vs_dagger} visualizes the corresponding
phase-space evolution on a representative rollout. Although BC-1200 delays
the onset of filamentation relative to BC-400, both offline policies
eventually develop substantial kinetic distortion and mode amplification.
DAgger-1 remains visibly closer to the expert evolution throughout the
rollout, exhibiting significantly less phase-space twisting and unstable mode
excitation at late times. Additional phase-space rollouts for different random initial phases are provided in Appendix~\ref{app:additional-phase-space-visualizations}.

\section{Conclusion}\label{sec:conclusion}

We studied imitation learning for stabilizing Vlasov--Poisson plasmas under partial observability. Our main result shows that small offline imitation error does not guarantee stable plasma rollouts: behavior cloning can induce exponentially growing instabilities through unresolved kinetic modes. In contrast, controlling the online imitation loss reduces this instability to polynomial growth over long horizons. Motivated by these theoretical insights, we introduced a practical DAgger-style algorithm and empirically validated its improved long-term stability over behavior cloning.

Our work opens several promising directions for future research. First, as we focus on the 1D torus setting, it would be interesting to extend our analysis and experiments to real-world 3-dimensional tokamak geometries. Second, an important direction is to study the convergence properties and sample complexity of the proposed DAgger-style algorithm. Finally, as the current study focuses on simulation environments, it would be valuable to investigate the practical implementation of online imitation learning for real-world plasma control systems.

\bibliographystyle{plain}
\bibliography{ref}

@article{Bishop1995,
  author  = {Bishop, Christopher M. and Haynes, Paul S. and Smith, Mike E. U. and Todd, Tom N. and Trotman, David L.},
  title   = {Real-Time Control of a Tokamak Plasma Using Neural Networks},
  journal = {Neural Computation},
  volume  = {7},
  number  = {1},
  pages   = {206--217},
  year    = {1995},
  doi     = {10.1162/neco.1995.7.1.206},
  note    = {\url{https://doi.org/10.1162/neco.1995.7.1.206}}
}

@article{KatesHarbeck2019,
  author  = {Kates-Harbeck, Julian and Svyatkovskiy, Alexey and Tang, William},
  title   = {Predicting Disruptive Instabilities in Controlled Fusion Plasmas through Deep Learning},
  journal = {Nature},
  volume  = {568},
  number  = {7753},
  pages   = {526--531},
  year    = {2019},
  doi     = {10.1038/s41586-019-1116-4},
  note    = {\url{https://doi.org/10.1038/s41586-019-1116-4}}
}

@article{Rea2019DIIID,
  author  = {Rea, C. and Montes, K. J. and Erickson, K. G. and Granetz, R. S. and Tinguely, R. A.},
  title   = {A Real-Time Machine Learning-Based Disruption Predictor in {DIII-D}},
  journal = {Nuclear Fusion},
  volume  = {59},
  number  = {9},
  pages   = {096016},
  year    = {2019},
  doi     = {10.1088/1741-4326/ab28bf},
  note    = {\url{https://doi.org/10.1088/1741-4326/ab28bf}}
}

@article{Montes2019Warnings,
  author  = {Montes, K. J. and Rea, C. and Granetz, R. S. and Tinguely, R. A. and Eidietis, N. and Meneghini, O. M. and Chen, D. L. and Shen, B. and Xiao, B. J. and Erickson, K. and Boyer, M. D.},
  title   = {Machine Learning for Disruption Warnings on {Alcator C-Mod}, {DIII-D}, and {EAST}},
  journal = {Nuclear Fusion},
  volume  = {59},
  number  = {9},
  pages   = {096015},
  year    = {2019},
  doi     = {10.1088/1741-4326/ab1df4},
  note    = {\url{https://doi.org/10.1088/1741-4326/ab1df4}}
}

@article{Akcay2021LockedMode,
  author  = {Ak{\c{c}}ay, Cihan and Finn, John M. and Brennan, Dylan P. and Burr, Thomas and K{\"u}rk{\c{c}}{\"u}o{\u{g}}lu, Do{\u{g}}a M.},
  title   = {Machine Learning Methods for Probabilistic Locked-Mode Predictors in Tokamak Plasmas},
  journal = {Physics of Plasmas},
  volume  = {28},
  number  = {8},
  pages   = {082106},
  year    = {2021},
  doi     = {10.1063/5.0053670},
  note    = {\url{https://doi.org/10.1063/5.0053670}}
}

@article{Fu2020MLControl,
  author  = {Fu, Yichen and Eldon, David and Erickson, Keith and Kleijwegt, Kornee and Lupin-Jimenez, Leonard and Boyer, Mark D. and Eidietis, Nick and Barbour, Nathaniel and Izacard, Olivier and Kolemen, Egemen},
  title   = {Machine Learning Control for Disruption and Tearing Mode Avoidance},
  journal = {Physics of Plasmas},
  volume  = {27},
  number  = {2},
  pages   = {022501},
  year    = {2020},
  doi     = {10.1063/1.5125581},
  note    = {\url{https://doi.org/10.1063/1.5125581}}
}

@article{Yang2022HL2A,
  author  = {Yang, Zongyu and Xia, Fan and Song, Xianming and Gao, Zhe and Li, Yixuan and Gong, Xinwen and Dong, Yunbo and Zhang, Yipo and Chen, Chengyuan and Luo, Cuiwen and Li, Bo and Zhu, Xiaobo and Ji, Xiaoquan and Li, Yonggao and Liu, Liang and Gao, Jinming and Liu, Yuhang},
  title   = {Real-Time Disruption Prediction in the Plasma Control System of {HL-2A} Based on Deep Learning},
  journal = {Fusion Engineering and Design},
  volume  = {182},
  pages   = {113223},
  year    = {2022},
  doi     = {10.1016/j.fusengdes.2022.113223},
  note    = {\url{https://doi.org/10.1016/j.fusengdes.2022.113223}}
}

@article{Murari2024ControlOriented,
  author  = {Murari, Andrea and Rossi, Riccardo and Craciunescu, Teddy and Vega, Jes{\'u}s and {JET Contributors} and others},
  title   = {A Control Oriented Strategy of Disruption Prediction to Avoid the Configuration Collapse of Tokamak Reactors},
  journal = {Nature Communications},
  volume  = {15},
  number  = {1},
  pages   = {2424},
  year    = {2024},
  doi     = {10.1038/s41467-024-46242-7},
  note    = {\url{https://doi.org/10.1038/s41467-024-46242-7}}
}

@article{Degrave2022,
  author  = {Degrave, Jonas and Felici, Federico and Buchli, Jonas and Neunert, Michael and Tracey, Brendan and Carpanese, Francesco and Ewalds, Timo and Hafner, Roland and Abdolmaleki, Abbas and de las Casas, Diego and Donner, Craig and Fritz, Leslie and Galperti, Cristian and Huber, Andrea and Keeling, James and Tsimpoukelli, Maria and Kay, Jackie and Merle, Antoine and Moret, Jean-Marc and Noury, Seb and Pesamosca, Federico and Pfau, David and Sauter, Olivier and Sommariva, Cristian and Coda, Stefano and Duval, Basil and Fasoli, Ambrogio and Kohli, Pushmeet and Kavukcuoglu, Koray and Hassabis, Demis and Riedmiller, Martin},
  title   = {Magnetic Control of Tokamak Plasmas through Deep Reinforcement Learning},
  journal = {Nature},
  volume  = {602},
  number  = {7897},
  pages   = {414--419},
  year    = {2022},
  doi     = {10.1038/s41586-021-04301-9},
  note    = {\url{https://doi.org/10.1038/s41586-021-04301-9}}
}

@article{Seo2024,
  author  = {Seo, Jaemin and Kim, SangKyeun and Jalalvand, Azarakhsh and Conlin, Rory and Rothstein, Andrew and Abbate, Joseph and Erickson, Keith and Wai, Josiah and Shousha, Ricardo and Kolemen, Egemen},
  title   = {Avoiding Fusion Plasma Tearing Instability with Deep Reinforcement Learning},
  journal = {Nature},
  volume  = {626},
  number  = {8000},
  pages   = {746--751},
  year    = {2024},
  doi     = {10.1038/s41586-024-07024-9},
  note    = {\url{https://doi.org/10.1038/s41586-024-07024-9}}
}

@inproceedings{Char2023,
  author    = {Char, Ian and Abbate, Joseph and Bardoczi, Laszlo and Boyer, Mark and Chung, Youngseog and Conlin, Rory and Erickson, Keith and Mehta, Viraj and Richner, Nathan and Kolemen, Egemen and Schneider, Jeff},
  title     = {Offline Model-Based Reinforcement Learning for Tokamak Control},
  booktitle = {Proceedings of The 5th Annual Learning for Dynamics and Control Conference},
  series    = {Proceedings of Machine Learning Research},
  volume    = {211},
  pages     = {1357--1372},
  year      = {2023},
  note      = {\url{https://proceedings.mlr.press/v211/char23a.html}}
}

@book{AriolaPironti2016,
  author    = {Ariola, Marco and Pironti, Alfredo},
  title     = {Magnetic Control of Tokamak Plasmas},
  publisher = {Springer},
  address   = {Cham},
  edition   = {2},
  year      = {2016},
  doi       = {10.1007/978-3-319-29890-0},
  note      = {\url{https://doi.org/10.1007/978-3-319-29890-0}}
}

@article{DeTommasi2019,
  author  = {De Tommasi, Gianmaria},
  title   = {Plasma Magnetic Control in Tokamak Devices},
  journal = {Journal of Fusion Energy},
  volume  = {38},
  number  = {3--4},
  pages   = {406--436},
  year    = {2019},
  doi     = {10.1007/s10894-018-0162-5},
  note    = {\url{https://doi.org/10.1007/s10894-018-0162-5}}
}

@phdthesis{Prechel2025,
  author = {Prechel, Garrett},
  title  = {Integrated Fast-Ion Diagnostics: Hardware Development, Synthetic Diagnostic Modeling and Phase Space Tomography},
  school = {University of California, Irvine},
  year   = {2025},
  type   = {{PhD} thesis},
  note   = {\url{https://escholarship.org/uc/item/2n6099z5}}
}

@article{Lao2005,
  author  = {Lao, L. L. and St. John, H. E. and Peng, Q. and Ferron, J. R. and Strait, E. J. and Taylor, T. S. and Meyer, W. H. and Zhang, C. and You, K. I.},
  title   = {{MHD} Equilibrium Reconstruction in the {DIII-D} Tokamak},
  journal = {Fusion Science and Technology},
  volume  = {48},
  number  = {2},
  pages   = {968--977},
  year    = {2005},
  doi     = {10.13182/FST48-968},
  note    = {\url{https://doi.org/10.13182/FST48-968}}
}

@article{Kinsey2022,
  author  = {Kinsey, J. E. and Lao, L. L. and Meneghini, O. and Candy, J. and Snyder, P. B. and Staebler, G. M.},
  title   = {Equilibrium Reconstruction of {DIII-D} Plasmas Using Predictive Modeling of the Pressure Profile},
  journal = {Physics of Plasmas},
  volume  = {29},
  number  = {6},
  pages   = {062502},
  year    = {2022},
  doi     = {10.1063/5.0078935},
  note    = {\url{https://doi.org/10.1063/5.0078935}}
}

@article{McClenaghan2024,
  author  = {McClenaghan, J. and Ak{\c{c}}ay, C. and Amara, T. B. and Sun, X. and Madireddy, S. and Lao, L. L. and Kruger, S. E. and Meneghini, O. M.},
  title   = {Augmenting Machine Learning of {G}rad--{S}hafranov Equilibrium Reconstruction with {G}reen's Functions},
  journal = {Physics of Plasmas},
  volume  = {31},
  number  = {8},
  pages   = {082507},
  year    = {2024},
  doi     = {10.1063/5.0213625},
  note    = {\url{https://doi.org/10.1063/5.0213625}}
}

@article{Morishita2024,
  author  = {Morishita, Yuya and Murakami, Sadayoshi and Kenmochi, Naoki and Funaba, Hisamichi and Yamada, Ichihiro and Mizuno, Yoshinori and Nagahara, Kazuki and Nuga, Hideo and Seki, Ryosuke and Yokoyama, Masayuki and Ueno, Genta and Osakabe, Masaki},
  title   = {First Application of Data Assimilation-Based Control to Fusion Plasma},
  journal = {Scientific Reports},
  volume  = {14},
  number  = {1},
  pages   = {137},
  year    = {2024},
  doi     = {10.1038/s41598-023-49432-3},
  note    = {\url{https://doi.org/10.1038/s41598-023-49432-3}}
}

@article{Landau1946,
  author  = {Landau, L. D.},
  title   = {On the Vibrations of the Electronic Plasma},
  journal = {Journal of Physics (USSR)},
  volume  = {10},
  pages   = {25--34},
  year    = {1946},
  note    = {\url{https://cds.cern.ch/record/437300}}
}

@article{Penrose1960,
  author  = {Penrose, Oliver},
  title   = {Electrostatic Instabilities of a Uniform Non-{M}axwellian Plasma},
  journal = {The Physics of Fluids},
  volume  = {3},
  number  = {2},
  pages   = {258--265},
  year    = {1960},
  doi     = {10.1063/1.1706024},
  note    = {\url{https://doi.org/10.1063/1.1706024}}
}

@book{Chen2016,
  author    = {Chen, Francis F.},
  title     = {Introduction to Plasma Physics and Controlled Fusion},
  edition   = {3},
  year      = {2016},
  publisher = {Springer},
  address   = {Cham},
  doi       = {10.1007/978-3-319-22309-4},
  note      = {\url{https://doi.org/10.1007/978-3-319-22309-4}}
}

@article{MouhotVillani2011,
  author  = {Mouhot, Cl{\'e}ment and Villani, C{\'e}dric},
  title   = {On {L}andau Damping},
  journal = {Acta Mathematica},
  volume  = {207},
  number  = {1},
  pages   = {29--201},
  year    = {2011},
  doi     = {10.1007/s11511-011-0068-9},
  note    = {\url{https://doi.org/10.1007/s11511-011-0068-9}}
}

@article{Glass2003,
  author  = {Glass, Olivier},
  title   = {On the Controllability of the {V}lasov--{P}oisson System},
  journal = {Journal of Differential Equations},
  volume  = {195},
  number  = {2},
  pages   = {332--379},
  year    = {2003},
  doi     = {10.1016/S0022-0396(03)00066-4},
  note    = {\url{https://doi.org/10.1016/S0022-0396(03)00066-4}}
}

@article{GlassHanKwan2012,
  author  = {Glass, Olivier and Han-Kwan, Daniel},
  title   = {On the Controllability of the {V}lasov--{P}oisson System in the Presence of External Force Fields},
  journal = {Journal of Differential Equations},
  volume  = {252},
  number  = {10},
  pages   = {5453--5491},
  year    = {2012},
  doi     = {10.1016/j.jde.2012.02.007},
  note    = {\url{https://doi.org/10.1016/j.jde.2012.02.007}}
}

@article{Einkemmer2024Constrained,
  author  = {Einkemmer, Lukas and Li, Qin and Wang, Li and Yang, Yunan},
  title   = {Suppressing Instability in a {V}lasov--{P}oisson System by an External Electric Field Through Constrained Optimization},
  journal = {Journal of Computational Physics},
  volume  = {498},
  pages   = {112662},
  year    = {2024},
  doi     = {10.1016/j.jcp.2023.112662},
  note    = {\url{https://doi.org/10.1016/j.jcp.2023.112662}}
}

@article{Bartsch2024,
  author  = {Bartsch, Jan and Knopf, Patrik and Scheurer, Stefania and Weber, J{\"o}rg},
  title   = {Controlling a {V}lasov--{P}oisson Plasma by a Particle-in-Cell Method Based on a {M}onte {C}arlo Framework},
  journal = {SIAM Journal on Control and Optimization},
  volume  = {62},
  number  = {4},
  pages   = {1977--2011},
  year    = {2024},
  doi     = {10.1137/23M1563852},
  note    = {\url{https://doi.org/10.1137/23M1563852}}
}

@article{EINKEMMER2025113904,
  author  = {Einkemmer, Lukas and Li, Qin and Mouhot, Cl{\'e}ment and Yue, Yukun},
  title   = {Control of Instability in a {V}lasov--{P}oisson System through an External Electric Field},
  journal = {Journal of Computational Physics},
  volume  = {530},
  pages   = {113904},
  year    = {2025},
  doi     = {10.1016/j.jcp.2025.113904},
  note    = {\url{https://doi.org/10.1016/j.jcp.2025.113904}}
}

@inproceedings{RossBagnell2010,
  author    = {Ross, Stephane and Bagnell, Drew},
  title     = {Efficient Reductions for Imitation Learning},
  booktitle = {Proceedings of the Thirteenth International Conference on Artificial Intelligence and Statistics},
  series    = {Proceedings of Machine Learning Research},
  volume    = {9},
  pages     = {661--668},
  year      = {2010},
  note      = {\url{https://proceedings.mlr.press/v9/ross10a.html}}
}

@inproceedings{RossGordonBagnell2011,
  author    = {Ross, Stephane and Gordon, Geoffrey and Bagnell, Drew},
  title     = {A Reduction of Imitation Learning and Structured Prediction to No-Regret Online Learning},
  booktitle = {Proceedings of the Fourteenth International Conference on Artificial Intelligence and Statistics},
  series    = {Proceedings of Machine Learning Research},
  volume    = {15},
  pages     = {627--635},
  year      = {2011},
  note      = {\url{https://proceedings.mlr.press/v15/ross11a.html}}
}

@article{RossBagnell2014,
  author  = {Ross, Stephane and Bagnell, J. Andrew},
  title   = {Reinforcement and Imitation Learning via Interactive No-Regret Learning},
  journal = {arXiv preprint arXiv:1406.5979},
  year    = {2014},
  note    = {\url{https://arxiv.org/abs/1406.5979}}
}

@inproceedings{Rajaraman2020,
  author    = {Rajaraman, Nived and Yang, Lin and Jiao, Jiantao and Ramchandran, Kannan},
  title     = {Toward the Fundamental Limits of Imitation Learning},
  booktitle = {Advances in Neural Information Processing Systems},
  volume    = {33},
  pages     = {2914--2924},
  year      = {2020},
  note      = {\url{https://proceedings.neurips.cc/paper/2020/hash/1e7875cf32d306989d80c14308f3a099-Abstract.html}}
}

@inproceedings{Foster2024,
  author    = {Foster, Dylan J. and Block, Adam and Misra, Dipendra},
  title     = {Is Behavior Cloning All You Need? Understanding Horizon in Imitation Learning},
  booktitle = {Advances in Neural Information Processing Systems},
  volume    = {37},
  pages     = {120602--120666},
  year      = {2024},
  doi       = {10.52202/079017-3834},
  note      = {\url{https://proceedings.neurips.cc/paper_files/paper/2024/hash/da84e39ae51fd26bb5110d9659c06e13-Abstract-Conference.html}}
}

@inproceedings{Simchowitz2025,
  author    = {Simchowitz, Max and Pfrommer, Daniel and Jadbabaie, Ali},
  title     = {The Pitfalls of Imitation Learning when Actions are Continuous},
  booktitle = {Proceedings of Thirty Eighth Conference on Learning Theory},
  series    = {Proceedings of Machine Learning Research},
  volume    = {291},
  pages     = {5248--5351},
  year      = {2025},
  note      = {\url{https://proceedings.mlr.press/v291/simchowitz25a.html}}
}

@article{xia2026provable,
  title   = {Provable Imitation Learning for Control of Instability in Partially-Observed {V}lasov--{P}oisson Equations},
  author  = {Xia, X. and Li, Q. and Mou, W.},
  journal = {arXiv preprint arXiv:2605.05081},
  year    = {2026},
  note = {\url{https://arxiv.org/abs/2605.05081}}
}

\newpage
\appendix

\section{Proof of results in \Cref{subsec:bc-limitation}}
\subsection{Proof of \Cref{thm:bc-limitation}}\label{app:bc-limitation}
We isolate the unstable linear response induced by the unobserved $k_2$ bias and show that nonlinear effects remain perturbative up to logarithmic time.

Throughout the proof, let $E[g]$ denote the electric field induced by a distribution $g$, and let $P_{k_2}$ and $P_{\ge k_2}$ denote Fourier projection onto modes $\pm k_2$ and modes $|k|\ge k_2$, respectively. We write
\[
E_t^\pi \coloneqq E[f_t^\pi],
\qquad
E_t^* \coloneqq E[f_t^*].
\]

We then analyze the nonlinear rollout. Let
\begin{align*}
    g_t \coloneqq f_t^\pi - f_t^*
\end{align*}
be the deviation from the expert trajectory. By construction of the BC policy, the deployed trajectory is driven by
\[
E_t^\pi+H_t^\pi
=
P_{\ge k_2}E_t^\pi
+
\epsilon_{\mathrm{tr}}\varphi_2 .
\]
Since the expert trajectory has no $k_2$ component, $g_t$ satisfies
\begin{align}
    \partial_t g_t
    =
    \mathcal{L}_{\mu}^{\ge k_2} g_t
    +
    \epsilon_{\mathrm{tr}}\varphi_2(x)\mu'(v)
    +
    \mathcal{N}(t,g_t),
    \label{eq:gt-dynamics}
\end{align}
where
\begin{align}
    \mathcal{L}_{\mu}^{\ge k_2} g
\coloneqq
-v\partial_x g + P_{\ge k_2}E[g]\,\mu'(v),
    \label{eq:restricted-linear-operator}
\end{align}
and the nonlinear remainder is
\begin{align}
    \mathcal{N}(t,g_t)
\coloneqq
\Big(P_{\ge k_2}E[g_t]
+
\epsilon_{\mathrm{tr}}\varphi_2\Big)
\partial_v(g_t+h_t).
    \label{eq:nonlinear-remainder}
\end{align}
Here
\[
h_t \coloneqq f_t^*-\mu
\]
is the expert perturbation. It remains supported on the $k_1$ mode and satisfies the Sobolev growth bound in \Cref{lem:expert-perturbation-growth}.

\begin{lemma}[Algebraic growth of the expert perturbation]
\label{lem:expert-perturbation-growth}
Fix a realization of
\[
    f_0(x,v)=\mu(v)\big(1+\zeta\cos(k_1x+\Theta)\big),
    \qquad |\zeta|\le \epsilon_0 .
\]
Then
\[
    h_t(x,v)
    =
    \zeta \mu(v)\cos(k_1(x-vt)+\Theta)
\]
satisfies
\begin{align}
    \|h_t\|_{\mathbb{H}^3_{x,v}}
    \le
    C_H \epsilon_0 (1+t^3),
    \label{eq:expert-growth}
\end{align}
where $C_H>0$ depends only on $\mu$, $k_1$, and the domain.
\end{lemma}
See \Cref{app:expert-perturbation-growth} for the proof of \Cref{lem:expert-perturbation-growth}. 

We first analyze the linear response.
\begin{lemma}[Linear response along the unstable mode]
\label{lem:linear-growth}Let $g_t^{(L)}$ solve
\begin{align}
    \partial_t g_t^{(L)}
    =
    \mathcal{L}_{\mu}^{\ge k_2}g_t^{(L)}
    +
    \varphi_2(x)\mu'(v),
    \qquad
    g_0^{(L)}=0.
    \label{eq:linear-response}
\end{align}
There exist constants $a_2,C_\perp,C_L>0$ such that
\begin{align}
    \|E[g_t^{(L)}]\|_{\ltwospace_x}
    \ge
    a_2 e^{\gamma_2 t} - C_\perp e^{\frac{\gamma_2}{2} t},
    \label{eq:linear-electric-norm-lower}
\end{align}
and
\begin{align}
    \|g_t^{(L)}\|_{\mathbb{H}^3_{x,v}} \le C_L e^{\gamma_2 t},
    \label{eq:linear-state-norm-upper}
\end{align}
\end{lemma}
See \Cref{app:linear-growth} for the proof of \Cref{lem:linear-growth}.\\
We then control the nonlinear deviation from the linear dynamics.
\begin{lemma}[Nonlinear remainder estimate]\label{lem:nonlinear-remainder}
Define
\(
R_t \coloneqq g_t-\epsilon_{\mathrm{tr}}g_t^{(L)}.
\)
Assume that, for all $s\in[0,t]$, the state deviation satisfies the bootstrap condition:
\begin{align}
    \|g_s\|_{\mathbb{H}^3_{x,v}}
    \le
    B\epsilon_{\mathrm{tr}} e^{\gamma_2 s}
    \label{eq:bootstrap-assumption}
\end{align}
 Then, there exist deterministic constants $C_1,C_2>0$  such that:
\begin{align}
    \|R_t\|_{\mathbb{H}^3_{x,v}}
    \le
    C_1 B^2 \epsilon_{\mathrm{tr}}^2 e^{2\gamma_2 t}
    +
    C_2 B \epsilon_{\mathrm{tr}}\epsilon_0 P(t)e^{\gamma_2 t}
    \label{eq:remainder-bound}
\end{align}
almost surely, where $P$ is a polynomial of degree at most four. Consequently,
\(
\|P_{k_2}E[R_t]\|_{\ltwospace_x}
\)
satisfies the same bound.
\end{lemma}
See \Cref{app:nonlinear-remainder} for the proof of \Cref{lem:nonlinear-remainder}.

We now prove \Cref{thm:bc-limitation}. Decompose
\begin{align}
    g_t = \epsilon_{\mathrm{tr}}g_t^{(L)} + R_t,
    \label{eq:linear-nonlinear-decomp}
\end{align}
By \Cref{lem:linear-growth},
\begin{align}
    \|g_t^{(L)}\|_{\mathbb{H}^3_{x,v}}
    \le C_L e^{\gamma_2 t},
    \label{eq:linear-state-upper-main}
\end{align}
and
\begin{align}
    \|E[g_t^{(L)}]\|_{\ltwospace_x}
    \ge
    a_2 e^{\gamma_2 t}
    -
    C_\perp e^{\frac{\gamma_2}{2}t}.
    \label{eq:linear-electric-lower-main}
\end{align}
Set \(B\coloneqq 2C_L\). Fix a threshold \(\delta_0>0\) and define
\begin{align}
    T^*
    \coloneqq
    \frac{1}{\gamma_2}
    \log\left(\frac{\delta_0}{a_2\epsilon_{\mathrm{tr}}}\right).
    \label{eq:Tstar-proof}
\end{align}
Then for all \(t\le T^*\),
\[
    \epsilon_{\mathrm{tr}}e^{\gamma_2 t}
    \le
    \frac{\delta_0}{a_2}.
\]
Define the maximal bootstrap time
\[
    T_{\max}
    \coloneqq
    \sup\Big\{
    t\in[0,T^*]:
    \|g_s\|_{\mathbb{H}^3_{x,v}}
    \le B\epsilon_{\mathrm{tr}}e^{\gamma_2s}
    \text{ for all }s\in[0,t]
    \Big\}.
\]
 Since $g_0=0$, continuity implies $T_{\max}>0$. We claim that
$T_{\max}=T^*$.

For any \(t<T_{\max}\), \Cref{lem:nonlinear-remainder} gives
\begin{align*}
    \|g_t\|_{\mathbb{H}^3_{x,v}}
    &\le
    \epsilon_{\mathrm{tr}}\|g_t^{(L)}\|_{\mathbb{H}^3_{x,v}}
    +
    \|R_t\|_{\mathbb{H}^3_{x,v}} \\
    &\le
    C_L\epsilon_{\mathrm{tr}}e^{\gamma_2t}
    +
    C_1B^2\epsilon_{\mathrm{tr}}^2e^{2\gamma_2t}
    +
    C_2B\epsilon_{\mathrm{tr}}\epsilon_0P(t)e^{\gamma_2t}.
\end{align*}
Using \(\epsilon_{\mathrm{tr}}e^{\gamma_2t}\le \delta_0/a_2\), we obtain
\begin{align*}
    \|g_t\|_{\mathbb{H}^3_{x,v}}
    \le
    \left(
    C_L
    +
    C_1B^2\frac{\delta_0}{a_2}
    +
    C_2B\epsilon_0P(T^*)
    \right)
    \epsilon_{\mathrm{tr}}e^{\gamma_2t}.
\end{align*}
Choose
\begin{align}
    \delta_0 &= \frac{a_2}{4 C_1 B^2} \min\left( C_L, a_2 \right), \label{eq:delta0-def} \\
    \epsilon_0 &= \frac{1}{4 C_2 B P(T^*)} \min\left( C_L, a_2 \right). \label{eq:eps0-def}
\end{align}
so that
\begin{align}
    C_1 B^2 \frac{\delta_0}{a_2} + C_2 B \epsilon_0 P(T^*) = \frac{1}{2} \min\left( C_L, a_2 \right).
    \label{eq:smallness-condition}
\end{align}
Using \(B=2C_L\), we obtain
\begin{align*}
    \|g_t\|_{\mathbb{H}^3_{x,v}} 
    &\le \left( C_L + \frac{1}{2} C_L \right) \epsilon_{\mathrm{tr}} e^{\gamma_2 t} 
    = \frac{3}{4} B \epsilon_{\mathrm{tr}} e^{\gamma_2 t}.
\end{align*}
By standard continuity arguments, we have $T_{\max} = T^*$, meaning the bootstrap assumption and Lemma~\ref{lem:nonlinear-remainder} hold securely on the entire window $[0, T^*]$.\\
We now estimate the electric field. Projecting onto the \(k_2\) mode yields $P_{k_2}E_t^\pi = \epsilon_{\mathrm{tr}}E[g_t^{(L)}] + P_{k_2}E[R_t]$.\\
By applying the reverse triangle inequality and the lower bound \eqref{eq:linear-electric-norm-lower}, we obtain:
\begin{align*}
    \|P_{k_2}E_t^\pi\|_{\ltwospace_x} 
    &\ge \epsilon_{\mathrm{tr}}\|E[g_t^{(L)}]\|_{\ltwospace_x} - \|P_{k_2}R_t\|_{\ltwospace_x} \\
    &\ge \left( a_2 \epsilon_{\mathrm{tr}}e^{\gamma_2 t} - C_\perp \epsilon_{\mathrm{tr}} e^{\frac{\gamma_2}{2}t} \right) - \|R_t\|_{\mathbb{H}^3_{x,v}}.
\end{align*}
By \eqref{eq:smallness-condition},
\[
\|R_t\|_{\mathbb{H}^3_{x,v}}
\le
\frac12 a_2\epsilon_{\mathrm{tr}}e^{\gamma_2 t}.
\] Thus, for all $t \in [0, T^*]$:
\begin{align*}
    \|E_{|k|=k_2}[f_t^\pi]\|_{\ltwospace_x} \ge \frac{1}{2} a_2 \epsilon_{\mathrm{tr}}e^{\gamma_2 t} - C_\perp \epsilon_{\mathrm{tr}} e^{\frac{\gamma_2}{2}t}.
\end{align*}
Since Fourier projections are orthogonal in \(\ltwospace_x\),
\[
    \|E[f_t^\pi]\|_{\ltwospace_x}
    \ge
    \|E_{|k|=k_2}[f_t^\pi]\|_{\ltwospace_x}.
\]
Evaluating at \(t=T^*\) and using
\(a_2\epsilon_{\mathrm{tr}}e^{\gamma_2T^*}=\delta_0\), we obtain
\begin{align*}
    \|E[f_{T^*}^\pi]\|_{\ltwospace_x}
    &\ge
    \frac{1}{2}\delta_0
    -
    C_\perp\epsilon_{\mathrm{tr}}e^{\frac{\gamma_2}{2}T^*} \\
    &=
    \frac{1}{2}\delta_0
    -
    C_\perp\sqrt{\frac{\delta_0}{a_2}}\,
    \epsilon_{\mathrm{tr}}^{1/2}
\end{align*}
For sufficiently small \(\epsilon_{\mathrm{tr}}\),
\begin{align*}
    \|E[f_{T^*}^\pi]\|_{\ltwospace_x} = \Omega(1).
\end{align*}
This completes the proof.
\subsection{Proof of \Cref{lem:expert-perturbation-growth}}\label{app:expert-perturbation-growth}
For any multi-index $\alpha=(\alpha_x,\alpha_v)$ with $|\alpha|\le 3$, spatial and velocity derivatives of $\cos(k_1(x-vt)+\Theta)$ generate at most polynomial factors in $t$. In particular, each $v$-derivative contributes a factor proportional to $t$, whereas $x$-derivatives only contribute powers of $k_1$. Hence derivatives of order at most three grow at most as $1+t^3$. Since $\mu$ is a Gaussian mixture, $\mu$ and its derivatives up to order three are rapidly decaying in $v$ and have finite $\mathbb{H}^3_{x,v}$ norms. Therefore,
\[
    \|h_t\|_{\mathbb{H}^3_{x,v}}
    \le
    C_H |\zeta|(1+t^3)
    \le
    C_H \epsilon_0(1+t^3).
\]
This proves the claim.
\subsection{Proof of \Cref{lem:linear-growth}}\label{app:linear-growth}
The operator $\mathcal{L}_{\mu}^{\ge k_2}$ preserves each spatial Fourier
mode. Since the initial data vanish and the forcing term is supported only
on the $k_2$ Fourier mode, the solution $g_t^{(L)}$ remains entirely
supported on the $k_2$ mode for all $t \ge 0$. We therefore analyze the dynamics on this invariant subspace.

Recall from Section~\ref{subsec:bc-limitation} that the mode $k_2$
lies in the Penrose-unstable regime, i.e., $k_2^2 < M$.
Consequently, on this invariant $k_2$ subspace, $\mathcal{L}_{\mu}^{\ge k_2}$
admits a real unstable eigenvalue $\gamma_2 > 0$. This eigenvalue coincides with the unique positive solution of the dispersion relation \eqref{eq:dispersion-complex} associated with $k_2$.

Let $\psi_2(x,v)$ and $\widetilde{\psi}_2(x,v)$ denote the corresponding
right and left eigenfunctions in $L^2_{x,v}$, satisfying
\[
    \mathcal{L}_{\mu}^{\ge k_2} \psi_2 = \gamma_2 \psi_2,
    \qquad
    (\mathcal{L}_{\mu}^{\ge k_2})^* \widetilde{\psi}_2 = \gamma_2 \widetilde{\psi}_2.
\]
We enforce the bi-orthogonality condition $\langle \psi_2, \widetilde{\psi}_2 \rangle_{\ltwospace_{x,v}} = 1$ to define the exact spectral projection. Since the eigenfunction $\psi_2$ induces a nontrivial electric field through the Poisson coupling, we scale both eigenfunctions to normalize it such that
\[
    \|E[\psi_2]\|_{\ltwospace_x} = 1.
\]
The projection of the forcing term onto this unstable eigendirection is given by:
\begin{align*}
    c_2
    \coloneqq
    \big\langle
        \varphi_2(x)\mu'(v),\widetilde{\psi}_2
    \big\rangle_{\ltwospace_{x,v}}.
\end{align*}
By adjusting the sign of $\psi_2$ and $\widetilde{\psi}_2$, we may assume $c_2>0$.
Define the unstable amplitude
\[
    A_2(t)
    \coloneqq
    \big\langle g_{t}^{(L)},\widetilde{\psi}_2
    \big\rangle_{\ltwospace_{x,v}}.
\]
Taking the inner product with $\widetilde{\psi}_2$ on both sides of \eqref{eq:linear-response} gives
\begin{align*}
    \frac{d}{dt}A_2(t)
    =
    \gamma_2 A_2(t) + c_2,
    \qquad
    A_2(0)=0.
\end{align*}
Therefore,
\begin{align*}
    A_2(t)
    =
    \frac{c_2}{\gamma_2}
    \big(e^{\gamma_2 t}-1\big).
\end{align*}
To isolate the dominant unstable contribution, we define the spectral projection operator $\Pi_2$ associated with the isolated eigenvalue $\gamma_2$ by
\[
    \Pi_2 g \coloneqq \big\langle g, \widetilde{\psi}_2 \big\rangle_{\ltwospace_{x,v}} \psi_2.
\]
By bi-orthogonality, $\Pi_2^2=\Pi_2$, and we decompose
\[
    g_t^{(L)} = \Pi_2 g_t^{(L)} + (I-\Pi_2)g_t^{(L)}.
\]
From the amplitude computation above,
\[
    \Pi_2 g_t^{(L)} = A_2(t)\psi_2,
    \qquad
    A_2(t)=\frac{c_2}{\gamma_2}(e^{\gamma_2 t}-1).
\]
Define the complementary component
\[
    g_t^\perp \coloneqq (I-\Pi_2)g_t^{(L)}.
\]
Since $\Pi_2$ commutes with the linear operator $\mathcal{L}_{\mu}^{\ge k_2}$, applying $(I-\Pi_2)$ to \eqref{eq:linear-response} yields:
\[
    \partial_t g_t^\perp
    =
    \mathcal{L}_{\mu}^{\ge k_2} g_t^\perp
    +
    (I-\Pi_2)\big(\varphi_2(x)\mu'(v)\big),
    \qquad g_0^\perp = 0.
\]
Let $\mathcal{L}^\perp$ denote the restriction of $\mathcal{L}_{\mu}^{\ge k_2}$ to the complementary invariant subspace $\mathrm{Im}(I-\Pi_2)$. Notice that the linearized Vlasov operator consists of a free transport part and a bounded perturbation. Since the spectral projection $\Pi_2$ exactly removes the unique unstable eigenvalue $\gamma_2$, the resolvent of $\mathcal{L}^\perp$ is uniformly bounded in any half-plane $\{\Re z \ge \beta\}$ for $\beta \in (0, \gamma_2)$. 

Because $\mathbb{H}^3_{x,v}$ is a Hilbert space, we invoke the Gearhart-Prüss theorem, which equates the semigroup growth bound with the resolvent bound. It immediately follows that there exists a constant $M_\beta > 0$ such that the generated semigroup satisfies:
\[
    \|e^{t\mathcal{L}^\perp}\|_{\mathcal{B}(\mathbb{H}^3_{x,v})}
    \le
    M_\beta e^{\beta t}.
\]
We take $\beta=\gamma_2/2$.
By Duhamel's formula,
\[
    g_t^\perp
    =
    \int_0^t
    e^{(t-s)\mathcal{L}^\perp}
    (I-\Pi_2)(\varphi_2\mu') \, ds,
\]
Since the Poisson operator is bounded from $\mathbb{H}^3_{x,v}$ to $L^2_x$, the semigroup bound implies that the macroscopic electric field of the complementary part has slower exponential growth:
\[
    \|E[g_t^\perp]\|_{\ltwospace_x}
    \le
    C_\perp e^{\beta t}.
\]
Therefore, using $\|E[\psi_2]\|_{\ltwospace_x}=1$,
\begin{align*}
    \|E[g_t^{(L)}]\|_{\ltwospace_x}
    &\ge
    |A_2(t)| - \|E[g_t^\perp]\|_{\ltwospace_x} \\
    &\ge
    \frac{c_2}{\gamma_2}(e^{\gamma_2 t}-1)
    -
    C_\perp e^{\beta t}.
\end{align*}
Adjusting constants and substituting $\beta=\gamma_2/2$ gives the desired lower bound: 
\[
    \|E[g_t^{(L)}]\|_{\ltwospace_x}
    \ge
    a_2 e^{\gamma_2 t} - C_\perp e^{\frac{\gamma_2}{2}t}.
\]
It remains to prove the corresponding Sobolev upper bound for the linear
response. Using the same decomposition,
\[
    g_{t}^{(L)}=A_2(t)\psi_2+g_t^\perp,
\]
we have
\[
    \|g_{t}^{(L)}\|_{\mathbb{H}^3_{x,v}}
    \le
    |A_2(t)|\|\psi_2\|_{\mathbb{H}^3_{x,v}}
    +
    \|g_t^\perp\|_{\mathbb{H}^3_{x,v}}.
\]
Since
\[
    A_2(t)=\frac{c_2}{\gamma_2}(e^{\gamma_2 t}-1),
\]
the first term is bounded by \(C e^{\gamma_2 t}\). On the complementary
subspace, the same Duhamel estimate in $\mathbb{H}^3_{x,v}$ gives
\[
    \|g_t^\perp\|_{\mathbb{H}^3_{x,v}}
    \le C e^{\frac{\gamma_2}{2}t}.
\]
Therefore, after increasing the constant if necessary, we have
\[
    \|g_{t}^{(L)}\|_{\mathbb{H}^3_{x,v}}
    \le
    C_L e^{\gamma_2 t}.
\]
This completes the proof.
\subsection{Proof of \Cref{lem:nonlinear-remainder}}\label{app:nonlinear-remainder}
Subtracting the scaled linear response equation $\epsilon_{\mathrm{tr}}\partial_t g_t^{(L)}$ from the exact deviation equation \eqref{eq:gt-dynamics}, the purely linear forcing term exactly cancels out. This reveals that the nonlinear remainder $R_t$ satisfies the unforced equation:
\begin{align}
    \partial_t R_t
    =
    \mathcal{L}_{\mu}^{\ge k_2}R_t
    +
    \mathcal{N}(t,g_t),
    \qquad
    R_0=0.
    \label{eq:R-dynamics}
\end{align}
Since $H^3 \hookrightarrow W^{1,\infty}$ in one spatial dimension, the Sobolev space $\mathbb{H}^3_{x,v}$ forms a Banach algebra. Combined with the boundedness of the Poisson operator from $\mathbb{H}^3_{x,v}$ to $\mathbb{H}^3_x$, the nonlinear remainder can be bounded by:
\begin{align*}
    \|\mathcal{N}(t,g_t)\|_{\mathbb{H}^3_{x,v}}
    &\le
    C_{\mathrm{alg}}
    \Big\|
        E_{|k|\ge k_2}[g_t]
        +
        \epsilon_{\mathrm{tr}}\varphi_2(x)
    \Big\|_{\mathbb{H}^3_x}
    \,
    \|g_t+h_t\|_{\mathbb{H}^3_{x,v}}                                                \\
    &\le
    \tilde{C}
    \big(
        \|g_t\|_{\mathbb{H}^3_{x,v}}
        +
        \epsilon_{\mathrm{tr}}
    \big)
    \big(
        \|g_t\|_{\mathbb{H}^3_{x,v}}
        +
        \|h_t\|_{\mathbb{H}^3_{x,v}}
    \big).
\end{align*}
We now explicitly insert the bootstrap assumption $\|g_t\|_{\mathbb{H}^3_{x,v}} \le B\epsilon_{\mathrm{tr}} e^{\gamma_2 t}$ and the algebraic growth bound of the expert perturbation $\|h_t\|_{\mathbb{H}^3_{x,v}} \le C_H \epsilon_0(1+t^3)$ from \Cref{lem:expert-perturbation-growth}. Assuming without loss of generality that the macroscopic scale $B \ge 1$, we have
\begin{align}
    \|\mathcal{N}(t,g_t)\|_{\mathbb{H}^3_{x,v}}
    &\le
    M_0 \big( B \epsilon_{\mathrm{tr}} e^{\gamma_2 t} \big) \big( B\epsilon_{\mathrm{tr}} e^{\gamma_2 t} + C_H \epsilon_0(1+t^3) \big) \notag \\
    &\le 
    M_1 B^2\epsilon_{\mathrm{tr}}^2 e^{2\gamma_2 t}
    +
    M_2 B \epsilon_{\mathrm{tr}}\epsilon_0(1+t^3)e^{\gamma_2 t},
    \label{eq:nonlinear-forcing-bound}
\end{align}
where $M_1$ and $M_2$ are universal constants independent of $B$, $\epsilon_{\mathrm{tr}}$, and $\epsilon_0$.\\
Applying Duhamel's formula to \eqref{eq:R-dynamics}, the remainder is given by $R_t = \int_0^t e^{(t-s)\mathcal{L}_{\mu}^{\ge k_2}} \mathcal{N}(s,g_s) \,ds$. As established in our semigroup analysis, the fundamental solution in the Sobolev space is strictly bounded by the dominant exponential growth $\|e^{\tau \mathcal{L}_{\mu}^{\ge k_2}} u\|_{\mathbb{H}^3_{x,v}} \le C_L e^{\gamma_2 \tau} \|u\|_{\mathbb{H}^3_{x,v}}$. Therefore:
\begin{align*}
    \|R_t\|_{\mathbb{H}^3_{x,v}} 
    &\le
    C_L
    \int_0^t
    e^{\gamma_2(t-s)}
    \|\mathcal{N}(s,g_s)\|_{\mathbb{H}^3_{x,v}}
    \,ds         \\
    &\le
    C_L M_1 B^2 \epsilon_{\mathrm{tr}}^2 \int_0^t e^{\gamma_2(t-s)} e^{2\gamma_2 s} \, ds 
    + C_L M_2 B \epsilon_{\mathrm{tr}} \epsilon_0 \int_0^t e^{\gamma_2(t-s)} (1+s^3)e^{\gamma_2 s} \, ds \\
    &\le
    C_1 B^2 \epsilon_{\mathrm{tr}}^2 e^{2\gamma_2 t}
    +
    C_2 B \epsilon_{\mathrm{tr}}\epsilon_0 P(t)e^{\gamma_2 t},
\end{align*}
where $P(t) = \int_0^t (1+s^3) ds = t + \frac{1}{4}t^4 = \mathcal{O}(1+t^4)$ is a polynomial of degree at most four, and $C_1$ and $C_2$ are deterministic constants independent of $B$, $\epsilon_{\mathrm{tr}}$, $\epsilon_0$, and $t$.\\ 
Finally, since the Poisson operator is a bounded linear map from $\mathbb{H}^3_{x,v}$ to $\ltwospace_x$, the projected electric field $\|P_{k_2}E[R_t]\|_{\ltwospace_x}$ trivially satisfies the same upper bound up to a multiplicative constant, which can be safely absorbed into $C_1$ and $C_2$. This completes the proof of the nonlinear remainder estimate.

\subsection{Proof of \Cref{prop:bc-lin-finite}}\label{app:bc-lin-finite}

Let
\begin{align*}
    F \coloneqq A + B K^*, \qquad
    \widehat{F} \coloneqq A + B \widehat{K}, \qquad
    \Delta_K \coloneqq \widehat{K} - K^*, \qquad
    \Delta_b \coloneqq \widehat{b} - b^*.
\end{align*}
All expectations in this proof are taken with respect to the random initialization $x_0 \sim \nu_0$. Once $x_0$ is fixed, both $x_t^*$ and $\widehat{x}_t$ are deterministic ODE trajectories.

For an affine policy $\widehat{u}(x) = \widehat{K}x + \widehat{b}$, the behavior cloning error along the expert trajectory is
\begin{align*}
    \varepsilon_{\mathrm{tr}}^2
    &= \frac{1}{T} \int_0^T \Exs \big[ \vecnorm{\Delta_K x_t^* + \Delta_b}{2}^2 \big] dt.
\end{align*}
By the definition of $\widebar{x}_T$, we have
$\frac{1}{T}\int_0^T \Exs[x_t^* - \widebar{x}_T]dt = 0$. Therefore, writing
$a_T \coloneqq \Delta_K \widebar{x}_T + \Delta_b$, the cross term vanishes and
\begin{align*}
    \varepsilon_{\mathrm{tr}}^2
    &= \mathrm{tr}\big(\Delta_K \Sigma_T \Delta_K^\top\big) + \vecnorm{a_T}{2}^2.
\end{align*}
The coverage assumption $\Sigma_T \succeq \lambda_0 I_d$ implies
\begin{align*}
    \vecnorm{\Delta_K}{F} \leq \frac{\varepsilon_{\mathrm{tr}}}{\sqrt{\lambda_0}},
    \qquad
    \opnorm{\Delta_K} \leq \frac{\varepsilon_{\mathrm{tr}}}{\sqrt{\lambda_0}},
    \qquad
    \vecnorm{a_T}{2} \leq \varepsilon_{\mathrm{tr}}.
\end{align*}
Choose
\begin{align*}
    c_1 \leq \frac{\rho_0 \sqrt{\lambda_0}}{4 \opnorm{P_0}\opnorm{B}}.
\end{align*}
If $\varepsilon_{\mathrm{tr}} < c_1$, then
\begin{align*}
    \widehat{F}^\top P_0 + P_0 \widehat{F}
    &\preceq F^\top P_0 + P_0 F
    + 2\opnorm{P_0}\opnorm{B}\opnorm{\Delta_K} I_d \\
    &\preceq - \frac{\rho_0}{2} I_d.
\end{align*}
Let $\kappa_0 \coloneqq \opnorm{P_0}\opnorm{P_0^{-1}}$ and $\bar{\rho} \coloneqq \rho_0/(2\opnorm{P_0})$. The preceding Lyapunov inequality gives, for all $z \in \real^d$ and $t \geq 0$,
\begin{align*}
    \vecnorm{e^{\widehat{F}t}z}{2}^2 \leq \kappa_0 e^{-\bar{\rho}t}\vecnorm{z}{2}^2.
\end{align*}
In particular, $\widehat{F}$ is invertible and
\begin{align*}
    \opnorm{\widehat{F}^{-1}}
    \leq \int_0^\infty \opnorm{e^{\widehat{F}s}}ds
    \leq \frac{2\sqrt{\kappa_0}}{\bar{\rho}}.
\end{align*}

The learned equilibrium is $\widehat{x}_\infty = -\widehat{F}^{-1}B\widehat{b}$. Since $F x_\infty^* + Bb^* = 0$, we have
\begin{align*}
    \widehat{F}(\widehat{x}_\infty - x_\infty^*)
    = B\big((K^*-\widehat{K})x_\infty^* + b^*-\widehat{b}\big).
\end{align*}
Consequently,
\begin{align*}
    \vecnorm{\widehat{x}_\infty - x_\infty^*}{2}
    &\leq \opnorm{\widehat{F}^{-1}}\opnorm{B}
    \left(\opnorm{\Delta_K}\vecnorm{x_\infty^*-\widebar{x}_T}{2} + \vecnorm{a_T}{2}\right) \\
    &\leq \frac{2\sqrt{\kappa_0}\opnorm{B}}{\bar{\rho}}
    \left(1 + \frac{\vecnorm{x_\infty^*-\widebar{x}_T}{2}}{\sqrt{\lambda_0}}\right)
    \varepsilon_{\mathrm{tr}}.
\end{align*}
It remains to note that the deterministic average $\widebar{x}_T$ is uniformly bounded around $x_\infty^*$ in terms of the second moment of $x_0$. Indeed, the expert Lyapunov condition gives
\begin{align*}
    \Exs\big[\vecnorm{x_t^*-x_\infty^*}{2}^2\big]
    \leq \kappa_0 e^{-\rho_0 t/\opnorm{P_0}}
    \Exs\big[\vecnorm{x_0-x_\infty^*}{2}^2\big],
\end{align*}
and hence, by Jensen's inequality,
\begin{align*}
    \vecnorm{x_\infty^*-\widebar{x}_T}{2}
    &\leq \frac{1}{T}\int_0^T \Exs\big[\vecnorm{x_t^*-x_\infty^*}{2}\big]dt \\
    &\leq \sqrt{\kappa_0 \Exs\big[\vecnorm{x_0-x_\infty^*}{2}^2\big]}.
\end{align*}
Thus there is a constant $C_{\infty}$ depending only on $(A,B,K^*,b^*,P_0,\rho_0,\lambda_0)$ and the distribution of $x_0$, but not on $T$, such that
\begin{align*}
    \vecnorm{\widehat{x}_\infty - x_\infty^*}{2}
    \leq C_{\infty}\varepsilon_{\mathrm{tr}}.
\end{align*}

Finally, since $\widehat{x}_t-\widehat{x}_\infty=e^{\widehat{F}t}(x_0-\widehat{x}_\infty)$,
\begin{align*}
    \Exs\big[\vecnorm{\widehat{x}_t - x_\infty^*}{2}^2\big]
    &\leq 2\Exs\big[\vecnorm{\widehat{x}_t-\widehat{x}_\infty}{2}^2\big]
    + 2\vecnorm{\widehat{x}_\infty-x_\infty^*}{2}^2\\
    &\leq 2\kappa_0 e^{-\bar{\rho}t}
    \Exs\big[\vecnorm{x_0-\widehat{x}_\infty}{2}^2\big]
    + 2C_{\infty}^2\varepsilon_{\mathrm{tr}}^2\\
    &\leq 4\kappa_0 e^{-\bar{\rho}t}
    \Exs\big[\vecnorm{x_0-x_\infty^*}{2}^2\big]
    + (4\kappa_0 + 2)C_{\infty}^2\varepsilon_{\mathrm{tr}}^2.
\end{align*}
Absorbing the fixed constants into $c_2$ proves the desired mean-square bound.
\section{Proof of results in \Cref{subsec:dagger-guarantees}}
\label{app:dagger-guarantees}
\subsection{Proof of \Cref{thm:stability-via-excess-risk}}\label{app:stability-via-excess-risk}
In this section, we prove the macroscopic stability of the closed-loop system under the learned policy $\pi$. The core mathematical challenge is that the online imitation loss $\mathcal{L}_{\mathrm{online}}(\pi)$ provides a bound on the $\ltwospace$ norm of the policy error integrated over time, whereas the nonlinear Vlasov--Poisson dynamics are highly sensitive to the $\mathbb{L}^\infty$ macroscopic density fluctuations. We bridge this gap using a continuous bootstrap argument and 1D Sobolev interpolation.

In the proof of \Cref{thm:stability-via-excess-risk}, we will cruicially use the following technical lemma, which controls the spatial density for a transport equation under a perturbation in the $v$--direction.
\begin{lemma}\label{lemma:transport-perturbed-rho-bound}
  Consider the perturbed transport equation:
  \begin{align}\label{eq:perturbed-transport}
    \partial_t \ffunc_t + v \pt_x \ffunc_t + \delta_t(x) \pt_v \ffunc_t = 0.
  \end{align}
   Suppose the initial condition $\ffunc_0$ satisfies Assumption~\ref{assume:initial-regularity} with parameters $\lambda_0, B_0, q$ and $\int_{\mathbb T\times\mathbb R} f_0\,dv\,dx=1$.
  Define $D(t) \coloneqq (\int_0^t \vecnorm{\delta_\tau(\cdot)}{\mathbb{L}^{\infty}(\torus)}^2\,d\tau)^{1/2}$. Then there exist constants $C_1,C_2$ depending only on $B_0, \lambda_0,q$ such that for all $t\ge0$,
  \begin{align*}
  \|\rho_t(\cdot)-1\|_{\mathbb{L}^\infty(\mathbb T)}
  \le C_1e^{-C_2t}
+
C_1 D(t) (1 + t)^2
+
C_1\,(1+D(t)^q t^{\frac{q}{2}}) D(t)^2t^3.
\end{align*}
moreover, if $D(s) \le 1$ for all $s \in [0,t]$, by adjusting constants, we could simplify the bound to
\begin{align*}
  \|\rho_t(\cdot)-1\|_{\mathbb{L}^\infty(\mathbb T)}
\le C_1e^{-C_2t}
+
C_1 D(t) (1 + t)^{\frac{q}{2}+3}.
\end{align*}
\end{lemma}
See \Cref{app:transport-perturbed-rho-bound} for the proof of this lemma.

Along the rollout induced by $\pi$, define the control error
\[
    \delta_t(x)
    \coloneqq
    \ctrlE^\pi_t[f^\pi](x)-\ctrlE^*_t[f^\pi](x),
    \qquad
    D(t)
    \coloneqq
    \left(
    \int_0^t \|\delta_s\|_{\mathbb{L}^\infty(\mathbb T)}^2\,ds
    \right)^{1/2}.
\]
By the definition of the online imitation loss,
\[
    \mathbb E\left[
    \int_0^T \|\delta_t\|_{\mathbb{L}^2(\mathbb T)}^2\,dt
    \right]
    \le \epsilon_{\mathrm{tr}}^2.
\]
Therefore, for any confidence parameter $\eta\in(0,1)$, with probability at
least $1-\eta$,
\[
    \int_0^T \|\delta_t\|_{\mathbb{L}^2(\mathbb T)}^2\,dt
    \le
    \epsilon^2,
    \qquad
    \epsilon\coloneqq \epsilon_{\mathrm{tr}}/\sqrt{\eta}.
\]
We work on this event throughout the proof.

\paragraph{Step 1: From $\mathbb{L}^2$ to $\mathbb{L}^\infty$ via Sobolev interpolation.}
By the one-dimensional Gagliardo--Nirenberg inequality,
\[
    \|\delta_s\|_{\mathbb{L}^\infty}^2
    \le
    C_{\mathrm{GN}}
    \|\delta_s\|_{\mathbb{L}^2}
    \|\partial_x\delta_s\|_{\mathbb{L}^2}.
\]
Thus,
\[
    D(t)^2
    \le
    C_{\mathrm{GN}}
    \int_0^t
    \|\delta_s\|_{\mathbb{L}^2}
    \|\partial_x\delta_s\|_{\mathbb{L}^2}\,ds.
\]

\paragraph{Step 2: Bootstrap derivative bound.}
Assume temporarily that \(D(s)\le 1\) for \(s\le t\). By
\Cref{lemma:transport-perturbed-rho-bound},
\[
    \|\rho_s^\pi-1\|_{\mathbb{L}^\infty}
    \le
    C e^{-cs}
    +
    C D(s)(1+s)^m,
    \qquad
    m\coloneqq q/2+3.
\]
Since
\[
    \|\partial_x\delta_s\|_{\mathbb{L}^2}
    \le
    \|\partial_x\ctrlE_s^\pi\|_{\mathbb{L}^2}
    +
    \|\partial_x\ctrlE_s^*\|_{\mathbb{L}^2}
    \le
    L_1+\|\rho_s^\pi-1\|_{\mathbb{L}^\infty},
\]
we obtain
\[
    \|\partial_x\delta_s\|_{\mathbb{L}^2}
    \le
    L + P(s)D(s),
    \qquad
    P(s)=C(1+s)^m,
\]
where \(L>0\) absorbs \(L_1\) and the exponentially decaying term. We now introduce the bootstrap interval
\[
    T_{\mathrm{boot}}
    \coloneqq
    \sup\Big\{
    t\in[0,T]:
    D(s)\le1
    \text{ for all } s\in[0,t]
    \Big\}.
\]
Since \(D(0)=0\) and \(D(t)\) is continuous, we have
\(T_{\mathrm{boot}}>0\).

For every \(t<T_{\mathrm{boot}}\), the bootstrap assumption
\(D(s)\le1\) holds on \([0,t]\), and therefore all preceding estimates
remain valid on this interval.

\paragraph{Step 3: The Algebraic Bootstrap Argument.}
For any \(t<T_{\mathrm{boot}}\), substituting the derivative bound gives
\[
D(t)^2
\le
C_{\mathrm{GN}}L\int_0^t\|\delta_s\|_{\mathbb{L}^2}\,ds
+
C_{\mathrm{GN}}\int_0^tP(s)\|\delta_s\|_{\mathbb{L}^2}D(s)\,ds.
\]
Since \(D(s)\le D(t)\) and we have the following bounds derived from the Cauthy--Schwarz inequality:
\[
\int_0^t\|\delta_s\|_{\mathbb{L}^2}\,ds\le \epsilon t^{1/2},
\qquad
\int_0^tP(s)\|\delta_s\|_{\mathbb{L}^2}\,ds
\le
\epsilon \widetilde P(t),
\]
where
\[
    \widetilde P(t)
    \coloneqq
    \left(\int_0^tP(s)^2\,ds\right)^{1/2}
    \lesssim
    (1+t)^{m+1/2},
\]
we obtain
\[
    D(t)^2
    \le
    C\epsilon t^{1/2}
    +
    C\epsilon \widetilde P(t)D(t).
\]
Solving this quadratic inequality for $D(t)$ gives
\[
    D(t)
    \le
    C\epsilon \widetilde P(t)
    +
    C(\epsilon t^{1/2})^{1/2}.
\]
Choose
\[
    T^*
    =
    c_*
    \min\left\{
    \epsilon^{-\frac{1}{m+1/2}},
    \epsilon^{-2}
    \right\},
\]
where \(c_*>0\) is sufficiently small depending only on the constants above.
Then for all \(t\le T^*\),
\[
    C\epsilon\widetilde P(t)
    +
    C(\epsilon t^{1/2})^{1/2}
    \le
    \frac12.
\]
Hence,
\[
    D(t)\le \frac12,
    \qquad
    \forall t<\min\{T_{\mathrm{boot}},T^*\}.
\]
If \(T_{\mathrm{boot}}<T^*\), continuity of \(D\) implies
\[
    D(T_{\mathrm{boot}})
    \le
    \frac12
    <
    1,
\]
contradicting the definition of \(T_{\mathrm{boot}}\).
Therefore,
\[
    T_{\mathrm{boot}}\ge T^*,
\]
and consequently
\[
    D(t)\le \frac12,
    \qquad
    \forall t\in[0,T^*].
\]

\paragraph{Step 4: Final macroscopic bounds on the density and electric field.}Using simplified bound in \Cref{lemma:transport-perturbed-rho-bound} again and the bound on \(D(t)\), for all
\(t\le T^*\), we have
\[
    \|\rho_t^\pi-1\|_{\mathbb{L}^\infty}
    \le
    Ce^{-ct}
    +
    C(1+t)^m
    \left[
        \epsilon(1+t)^{m+1/2}
        +
        \epsilon^{1/2}(1+t)^{1/4}
    \right].
\]
Thus
\[
    \|\rho_t^\pi-1\|_{\mathbb{L}^\infty}
    \le
    Ce^{-ct}
    +
    C\epsilon(1+t)^{2m+1/2}
    +
    C\epsilon^{1/2}(1+t)^{m+1/4}.
\]
By the one-dimensional Poisson estimate,
\[
    \|E_t^\pi\|_{\mathbb{L}^2}
    \le
    C_P\|\rho_t^\pi-1\|_{\mathbb{L}^\infty}.
\]
Therefore,
\[
    \|E_t^\pi\|_{\mathbb{L}^2}
    \le
    C e^{-ct}
    +
    C\epsilon(1+t)^{2m+1/2}
    +
    C\epsilon^{1/2}(1+t)^{m+1/4},
    \qquad
    t\le T^*.
\]
Since the above bound holds for any $q\ge 2$, by choosing $q = 2$,using the inequality $(1+t)^{m+1/4} \le (1+t)^{2m+1/2}$ for $t\ge 0$ and the fact that $\epsilon^{1/2} \le \epsilon$ for $\epsilon \le 1$, we can further simplify the bound to
\[
    \|E_t^\pi\|_{\mathbb{L}^2}
    \le
    C e^{-ct}
    +
    C\epsilon(1+t)^9,
    \qquad
    t\le T^*.
\]
\subsection{Proof of \Cref{lemma:transport-perturbed-rho-bound}}\label{app:transport-perturbed-rho-bound}
For any $(t,x,v)$, let $(X(s),V(s))_{0\le s\le t}$ be the backward characteristic:
\[
\dot X(s)=V(s),\qquad \dot V(s)=\delta_s(X(s)),\qquad X(t)=x,\ V(t)=v.
\]
Define the free-streaming path $X_{\rm f}(s):=x-(t-s)v$ and the offsets
\[
\Delta v:=V(0)-v,\qquad \Delta x:=X(0)-X_{\rm f}(0)=X(0)-(x-tv).
\]
By Cauchy--Schwarz,
\[
\int_0^t \|\delta_s\|_{\mathbb{L}^\infty(\T)}\,ds
\le \Big(\int_0^t \|\delta_s\|_{\mathbb{L}^\infty(\T)}^2\,ds\Big)^{1/2}\, t^{1/2}
= D(t)\,t^{1/2}.
\]
Integrating $\dot V=\delta_s(X(s))$ gives
\[
\Delta v=\int_0^t \delta_{s}(X(s))\,ds,
\qquad\Rightarrow\qquad
|\Delta v|\le \int_0^t \vecnorm{\delta_{s}}{\mathbb{L}^\infty(\T)}\,ds \le D(t)\,t^{1/2}.
\]
Similarly, integrating $\dot X=V$ gives
\[
\Delta x=X(0)-(x-tv)=\int_0^t (V(\tau)-v)\,d\tau
=\int_0^t\int_\tau^t \delta_s(X(s))\,ds\,d\tau
=\int_0^t (t - s) \,\delta_s(X(s))\,ds,
\]
hence by Cauchy--Schwarz, we have
\[
|\Delta x|
\le \int_0^t (t-s)\|\delta_s\|_{\mathbb{L}^\infty(\T)}\,ds
\le \Big(\int_0^t (t-s)^2 ds\Big)^{1/2}\Big(\int_0^t \|\delta_s\|_{\mathbb{L}^\infty(\T)}^2 ds\Big)^{1/2}
= \frac{t^{3/2}}{\sqrt3}\,D(t).
\]
In particular,
\begin{equation}\label{eq:dx-dv-bounds}
|\Delta v|\le D(t)t^{1/2},\qquad |\Delta x|\le D(t)t^{3/2}.
\end{equation}
Along characteristics the solution is constant, so
\begin{equation}\label{eq:repr-f}
f_t(x,v)=f_0\big(X(0),V(0)\big)=f_0\big(x-tv+\Delta x,\ v+\Delta v\big).
\end{equation}
Assumption~\ref{assume:initial-regularity} states that
\[
\|f_0\|_{\mathcal E_q(\lambda_0)}=\sum_{m=0}^{\infty}\frac{\lambda_0^m}{m!}\|f_0\|_{\mathcal C^m_q}\le B_0,
\qquad
\|f_0\|_{\mathcal C^m_q}
:=\sup_{0\le a+b\le m}\ \sup_{(x,v)\in\mathbb T\times\mathbb R}(1+|v|)^q\,|\partial_x^a\partial_v^b f_0(x,v)|.
\]
Therefore, for each $m\ge 0$,
\begin{equation}\label{eq:Cm-bound}
\|f_0\|_{\mathcal C^m_q}\le B_0\,\frac{m!}{\lambda_0^m},
\end{equation}
because all terms in the series are nonnegative.
Consequently, for any integers $a,b\ge0$,
\begin{equation}\label{eq:L1-derivative}
\sup_{x\in\mathbb T}\int_{\mathbb R}\big|\partial_x^a\partial_v^b f_0(x,v)\big|\,dv
\le
\|f_0\|_{\mathcal C^{a+b}_q}\int_{\mathbb R}(1+|v|)^{-q}\,dv
\le
C_\ast\,B_0\,\frac{(a+b)!}{\lambda_0^{a+b}},
\end{equation}
where $C_\ast:=\int_{\mathbb R}(1+|v|)^{-q}\,dv<\infty$ is an absolute constant (since $q \ge 2 > 1$).\\
Define the free-streaming density
\[
\rho^{\rm fs}_t(x):=\int_{\mathbb R} f_0(x-tv,v)\,dv.
\]
Then, by adding and subtracting $\rho^{\rm fs}$, we obtain
\begin{equation}\label{eq:split}
|\rho_t(x)-1|
\le
|\rho^{\rm fs}_t(x)-1|
+
|\rho_t(x)-\rho^{\rm fs}_t(x)|.
\end{equation}
We start by estimating the first term in \eqref{eq:split}.
Define
\[
\rho_0(x):=\int_{\real} f_0(x,v)\,dv,
\qquad
\widehat{\rho_0}(k):=\int_{\torus}\rho_0(x)\,e^{-2\pi i k x}\,dx,
\qquad k\in\mathbb Z,
\]
so that $\widehat{\rho_0}(0)=\int_{\torus\times\real} f_0\,dv\,dx=1$.
For $k\neq 0$, define the mixed Fourier transform of $f_0$ by
\[
\widehat f_0(k,\eta)
:=\int_{\torus\times\real} f_0(x,v)\,e^{-2\pi i(kx+\eta v)}\,dx\,dv,
\qquad \eta\in\real,
\]
so that $\widehat{\rho_0}(k)=\widehat f_0(k,0)$.
Under free streaming,
\[
\rho^{\mathrm{fs}}_t(x)=\int_{\real} f_0(x-tv,v)\,dv,
\]
and for each $k\neq 0$,
\[
\widehat{\rho^{\mathrm{fs}}_t}(k)
=\int_{\torus\times\real} f_0(y,v)\,e^{-2\pi i(k y + k t v)}\,dy\,dv
=\widehat f_0(k,kt).
\]
Hence, using the Fourier series on $\torus$ and $\widehat{\rho^{\mathrm{fs}}_t}(0)=1$,
\begin{equation}\label{eq:rho-fs-fourier-sum}
\rho^{\mathrm{fs}}_t(x)-1
=\sum_{k\in\mathbb Z\setminus\{0\}}\widehat f_0(k,kt)\,e^{2\pi i k x},
\qquad
\|\rho^{\mathrm{fs}}_t-1\|_{L^\infty(\torus)}
\le \sum_{k\neq 0}\big|\widehat f_0(k,kt)\big|.
\end{equation}
We next bound $\widehat f_0(k,\eta)$ using Assumption~\ref{assume:initial-regularity}.
For any $m\in\mathbb N$, integrating by parts $m$ times in $v$ yields
\[
(2\pi i\eta)^m\widehat f_0(k,\eta)
=\int_{\torus\times\real}\partial_v^m f_0(x,v)\,e^{-2\pi i(kx+\eta v)}\,dx\,dv,
\]
hence
\[
|\eta|^m\,|\widehat f_0(k,\eta)|
\le (2\pi)^{-m}\|\partial_v^m f_0\|_{\mathbb{L}^1(\torus\times\real)}.
\]
Using the velocity weight in $\|\cdot\|_{\mathcal C^m_q}$ and
$\int_{\real}(1+|v|)^{-q}dv<\infty$ (since $q \ge 2 > 1$), we have
\[
\|\partial_v^m f_0\|_{\mathbb{L}^1(\torus\times\real)}
\le C_\ast\,\|f_0\|_{\mathcal C^m_q},
\]
where $C_\ast>0$ is universal.
By \eqref{eq:Cm-bound}, $\|f_0\|_{\mathcal C^m}\le B_0\,m!\,\lambda_0^{-m}$, hence
\[
|\eta|^m\,|\widehat f_0(k,\eta)|
\le C_\ast B_0\,m!\,(2\pi\lambda_0)^{-m},
\qquad \forall m\in\mathbb N.
\]
Fix $\delta = \pi\lambda_0$. Multiplying by $\delta^m/m!$ and summing over $m\ge0$ yields
\[
e^{\delta|\eta|}\,|\widehat f_0(k,\eta)|
\le C_\ast B_0\sum_{m=0}^\infty\Big(\frac{\delta}{2\pi\lambda_0}\Big)^m
=\frac{C_\ast B_0}{1-\delta/(2\pi\lambda_0)} = 2C_\ast B_0.
\]
Consequently, there exist constants $C_0,c_0>0$ depending only on $(\lambda_0,B_0)$ such that
\[
|\widehat f_0(k,\eta)|\le C_0\,e^{-c_0|\eta|},
\qquad \forall k\in\mathbb Z,\ \eta\in\mathbb R.
\]
Applying this bound with $\eta=kt$ in \eqref{eq:rho-fs-fourier-sum} gives
\[
\|\rho^{\mathrm{fs}}_t-1\|_{\mathbb{L}^\infty(\torus)}
\le C_0\sum_{k\neq 0}e^{-c_0|k|t}
=2C_0\sum_{k=1}^\infty e^{-c_0 k t}
=\frac{2C_0 e^{-c_0 t}}{1-e^{-c_0 t}}.
\]
For $t\ge 1$,
\[
\frac{2C_0 e^{-c_0 t}}{1-e^{-c_0 t}}
\le C_1 e^{-c_0 t},
\]
with $C_1:=\frac{2C_0 }{1 - e^{-c_0}}$.
For $t\in[0,1]$, we simply use
\[
\|\rho_t^{\mathrm{fs}}\|_{L^\infty(\T)}
\le \sup_{x\in\T}\int_{\R}|f_0(x-tv,v)|\,dv
= \sup_{x\in\T}\int_{\R}|f_0(x,v)|\,dv
\le C(\lambda_0,B_0),
\]
and hence $\|\rho_t^{\mathrm{fs}}-1\|_{\mathbb{L}^\infty(\T)}\le C(\lambda_0,B_0)$ on $[0,1]$. Combining the two cases yields
\begin{align}\label{eq:rho-fs-bound}
   \|\rho_t^{\mathrm{fs}}-1\|_{\mathbb{L}^\infty(\torus)}\le C_1 e^{- c_2 t}
\end{align}
 for all $t\ge0$ for some constants $C_1,c_2>0$ depending only on $(\lambda_0,B_0)$.
It remains to bound the second term in \eqref{eq:split} uniformly in $x$:
\[
\rho_t(x)-\rho_t^{\rm fs}(x)
=
\int_{\mathbb R}
\Big(
f_0(x-tv+\Delta x,\,v+\Delta v)-f_0(x-tv,\,v)
\Big)\,dv.
\]
Fix $(t,x)$ and for each $v\in\mathbb R$ denote the base point $(x_0,v_0):=(x-tv,v)$.
Applying the multivariate Taylor formula for $f_0$ at $(x_0,v_0)$ to first order yields
\begin{align}
f_0(x_0+\Delta x,\,v_0+\Delta v)
&= f_0(x_0,v_0)
+ \partial_x f_0(x_0,v_0)\,\Delta x
+ \partial_v f_0(x_0,v_0)\,\Delta v
+ R_1(t,x,v),
\label{eq:taylor-1st}
\end{align}
where the remainder admits the integral representation
\begin{equation}\label{eq:R1-int}
R_1(t,x,v)
=
\sum_{a+b=2}\frac{1}{a!\,b!}\,(\Delta x)^a(\Delta v)^b
\int_0^1 (1-s)\,
\partial_x^a\partial_v^b f_0(x_0+s\Delta x,\ v_0+s\Delta v)\,ds .
\end{equation}
Subtracting $f_0(x_0,v_0)$, integrating in $v$, and taking absolute values yield
\begin{align}\label{eq:rho-pert-1st}
|\rho_t(x)-\rho_t^{\rm fs}(x)|
&\le
\int_{\mathbb R} |\Delta x|\,|\partial_x f_0(x-tv,v)|\,dv
\;+\;
\int_{\mathbb R}|\Delta v|\,|\partial_v f_0(x-tv,v)|\,dv
\;+\;
\int_{\mathbb R}|R_1(t,x,v)|\,dv .
\end{align}
As for the linear term, applying the elementary inequality $\int |h(v)|\, |g(v)|\,dv\le \|h\|_{\mathbb{L}^\infty_v}\int |g(v)|\,dv$ and \eqref{eq:dx-dv-bounds}, we obtain
\begin{align}\label{eq:first-order-bound}
\int_{\mathbb R}|\Delta x(t,x,v)|\,|\partial_x f_0(x-tv,v)|\,dv
&\le \|\Delta x(t,x,\cdot)\|_{\mathbb{L}^\infty_v}\int_{\mathbb R}|\partial_x f_0(x-tv,v)|\,dv
\\
&\le D(t)t^{3/2} \int_{\mathbb R}|\partial_x f_0(x-tv,v)|\,dv,
\nonumber\\
\int_{\mathbb R}|\Delta v(t,x,v)|\,|\partial_v f_0(x-tv,v)|\,dv
&\le \|\Delta v(t,x,\cdot)\|_{\mathbb{L}^\infty_v}\int_{\mathbb R}|\partial_v f_0(x-tv,v)|\,dv
\\
&\le D(t)t^{1/2} \int_{\mathbb R}|\partial_v f_0(x-tv,v)|\,dv.
\nonumber
\end{align}
Using the $\mathbb{L}^1_v$ derivative bound \eqref{eq:L1-derivative} with $(a,b)=(1,0)$ and $(0,1)$ gives
\begin{equation}\label{eq:first-order-final}
\int_{\mathbb R}
|\Delta x|\,|\partial_x f_0(x-tv,v)|\,dv
+
\int_{\mathbb R}
|\Delta v|\,|\partial_v f_0(x-tv,v)|\,dv
\le
C(\lambda_0,B_0)\Big[D(t)t^{1/2} + D(t)t^{3/2}\Big].
\end{equation}
As for the remainder $R_1$, we bound it pointwise using \eqref{eq:Cm-bound} with $m = 2$.
Recall
\[
R_1(t,x,v)=\sum_{a+b=2}\frac{1}{a!\,b!}(\Delta x)^a(\Delta v)^b
\int_0^1(1-s)\,
\partial_x^a\partial_v^b f_0(x_0+s\Delta x,\ v_0+s\Delta v)\,ds,
\qquad (x_0,v_0)=(x-tv,v).
\]
Taking absolute values, integrating in $v$, and using $(1-s)\le 1$ yield
\begin{align}
\int_{\mathbb R}|R_1(t,x,v)|\,dv
&\le
\sum_{a+b=2}\frac{1}{a!\,b!}
\int_{\mathbb R}|\Delta x(t,x,v)|^a|\Delta v(t,x,v)|^b
\int_0^1
\big|
\partial_x^a\partial_v^b f_0(x_0+s\Delta x,\ v_0+s\Delta v) 
\big|\,ds\,dv . \nonumber\\
&\le C(\lambda_0,B_0)\int_{\mathbb R}(|\Delta x|^2 + |\Delta x| |\Delta v| + |\Delta v|^2)\int_0^1 (1+|v + s \Delta v|)^{-3} ds\,dv. \label{eq:R1-start}
\end{align}
For each fixed $v$ we have
\[
1+|v+s\Delta v|\ge \frac{1+|v|}{1+|s\Delta v|}\qquad\Rightarrow\qquad
(1+|v+s\Delta v|)^{-q}\le (1+|s\Delta v|)^q(1+|v|)^{-q}\le (1+|\Delta v|)^q(1+|v|)^{-q}.
\]
Plugging this into \eqref{eq:R1-start} and using $\int_0^1 ds=1$ gives
\begin{align}\label{eq:R1-mid}
\int_{\mathbb R}|R_1(t,x,v)|\,dv
&\le
C(\lambda_0,B_0)\int_{\mathbb R}(|\Delta x|^2 + |\Delta x| |\Delta v| + |\Delta v|^2)(1+|\Delta v|)^{q}(1+|v|)^{-q} \,dv.
\end{align}
Now use \eqref{eq:dx-dv-bounds} and $\int_{\R} (1 + |v|)^{-q} dv <\infty$ to bound the integral, we have:
\begin{align}\label{eq:R1-final}
\int_{\mathbb R}|R_1(t,x,v)|\,dv \le C(\lambda_0,B_0)(1 + D(t)t^{1/2})^q D(t)^2 t^3.
\end{align}
Combining \eqref{eq:first-order-final} and \eqref{eq:R1-final} with \eqref{eq:rho-pert-1st} gives
\begin{equation}\label{eq:rho-pert-final}
\|\rho_t(\cdot)-\rho_t^{\rm fs}(\cdot)\|_{\mathbb{L}^\infty(\mathbb T)}
\le
C(\lambda_0,B_0)\Big[D(t)t^{1/2} + D(t)t^{3/2}\Big]
+
C(\lambda_0,B_0)\,(1+D(t)t^{1/2})^qD(t)^2t^3.
\end{equation}
Finally, recalling \eqref{eq:split} and \eqref{eq:rho-fs-bound}, we conclude
\[
\|\rho_t(\cdot)-1\|_{\mathbb{L}^\infty(\mathbb T)}
\le
C(\lambda_0,B_0)e^{-c' t}
+
C(\lambda_0,B_0)\Big[D(t)t^{1/2} + D(t)t^{3/2}\Big]
+
C(\lambda_0,B_0)\,(1+D(t)t^{1/2})^qD(t)^2t^3.
\]
Finally, using $t^{1/2}\le 1+t$ and $t^{3/2}\le 1+t^2$ for all $t\ge0$, we could rewrite the bound in terms of polynomial expressions in $t$ as stated in the lemma, which completes the proof.

\section{Experimental Reproducibility Details}
\label{app:exp-reproducibility}

This appendix summarizes the implementation details required to reproduce the experiments in \Cref{sec:simulations}, including the numerical solver, training setup, DAgger fine-tuning protocol, and evaluation procedure. Source code is available from the authors upon reasonable request.

\subsection{Simulation Setup}
\label{app:solver-details}

All experiments are conducted on the one-dimensional Vlasov--Poisson system
described in the main text. We use a semi-Lagrangian solver implemented in
PyTorch with spatial and velocity resolutions
\[
n_x = 1024,
\qquad
n_v = 1024,
\]
and time step
\[
\Delta t = 0.02.
\]

The spatial domain is periodic with length
\[
L = 10\pi,
\]
and the velocity domain is truncated to
\[
v \in [-6,6].
\]

The equilibrium distribution is the symmetric two-stream state with
parameters
\[
(\bar v,\sigma) = (2.4,1.0).
\]
All rollout evaluations use horizon
\[
T = 80.
\]

All trajectories are initialized from the dual-mode perturbation
\[
f_0(x,v)
=
\mu(v)
\Big(
1
+
\varepsilon\cos(k_1x+\phi_1)
+
\varepsilon\cos(k_2x+\phi_2)
\Big),
\]
with
\[
k_1 = 1,
\qquad
k_2 = 5,
\qquad
\varepsilon = 0.05,
\]
and phases $(\phi_1,\phi_2)$ sampled uniformly from $[0,2\pi)$.

For rollout evaluation, the controller observes a spatially subsampled
density history with stride $256$. The first $50$ time steps are treated as
a warm-up period during which control is suppressed or replaced by the
corresponding baseline rule.


\subsection{Policy Training}
\label{app:training-details}

The offline behavior cloning datasets consist of expert-generated rollout
trajectories stored trajectory-wise on disk. Each trajectory contains
subsampled density observations and Fourier representations of the target
electric field.

The reported policies are:
\begin{itemize}
    \item \textbf{BC-400:} offline behavior cloning trained on $400$ expert trajectories.
    \item \textbf{BC-1200:} offline behavior cloning trained on $1200$ expert trajectories.
    \item \textbf{DAgger-1:} initialized from BC-400 and fine-tuned using one round of aggregated on-policy data.
\end{itemize}

For all offline training runs, trajectories are randomly partitioned into
train/validation/test splits using ratios
\[
80\% / 10\% / 10\%,
\]
with random seed $42$.

The policy input consists of a short temporal history of subsampled density
observations together with a first-order temporal-difference channel. The
policy output is a truncated Fourier representation of the electric field.

All reported controllers use the same temporal convolutional network with
multi-head self-attention. The architecture consists of causal residual TCN
blocks followed by an attention layer and a two-layer MLP prediction head.
The main hyperparameters are summarized below:
\begin{center}
\begin{tabular}{lc}
\toprule
Parameter & Value \\
\midrule
TCN channel width & 128 \\
Kernel size & 5 \\
Dropout & 0.1 \\
Attention heads & 4 \\
Dilation schedule & $(1,2,4,8)$ \\
\bottomrule
\end{tabular}
\end{center}

All policies are trained using AdamW. The main optimization hyperparameters
are:
\begin{center}
\begin{tabular}{lc}
\toprule
Parameter & Value \\
\midrule
Batch size & 1024 \\
Initial learning rate & $10^{-3}$ \\
Weight decay & $10^{-4}$ \\
Training epochs & 100 \\
\bottomrule
\end{tabular}
\end{center}

The training objective is a robust hybrid Fourier loss defined on the
normalized Fourier targets. Let $\hat y$ and $y$ denote the predicted and
target normalized Fourier coefficients. The base discrepancy is the
coefficientwise $\ell_2$ error. This error is then weighted in two ways: by
a static mode-dependent factor proportional to $1/k$, and by a dynamic
inverse-magnitude factor proportional to
\[
\max(|y|,\tau)^{-1},
\]
where $\tau$ is a per-channel stabilization threshold computed as the
empirical $20$th percentile of the absolute normalized training targets.
The final loss is a weighted sum of an absolute weighted term and a relative
weighted term, with mixing coefficients $\alpha=1.0$ and $\beta=0.3$.

Inputs and Fourier targets are normalized using global statistics computed
from the training split. The same normalization statistics are reused during
DAgger fine-tuning.


\subsection{DAgger Fine-Tuning}
\label{app:dagger-details}

DAgger-1 is initialized from the BC-400 checkpoint and fine-tuned using a
mixture of offline expert data and newly aggregated on-policy trajectories.

During aggregation rollouts, the controller action is applied with
probability
\[
\alpha = 0.6,
\]
while the expert controller is used otherwise. A total of $800$ aggregated
trajectories are collected for the reported DAgger-1 experiment.

To avoid contaminating the aggregated dataset with clearly divergent late-time rollouts, trajectories are truncated whenever the electric energy exceeds $10^{-4}$ after time $t=30$. This heuristic threshold acts as a safety filter, ensuring that the fine-tuning process focuses on stabilizing borderline recoverable regimes rather than attempting to fit macroscopic numerical blow-ups.

Fine-tuning reuses the same architecture and normalization statistics as the
original BC model. Aggregated trajectories are concatenated directly with the
offline dataset without additional upsampling. The fine-tuning optimizer is
AdamW with learning rate
\[
10^{-4}.
\]


\subsection{Evaluation Protocol and Figure Construction}
\label{app:evaluation-protocol}

The electric-energy plots in \Cref{sec:simulations} report the pointwise empirical mean and a $\pm 1$ standard deviation ($\pm 1\sigma$) shaded region. This statistical variability captures the uncertainty arising from the randomized initial phase-space perturbations across $10$ independent rollout trajectories with uniformly sampled random phases $(\phi_1,\phi_2)$. The same rollout
initializations are shared across all compared controllers to ensure a paired
comparison.

Phase-space visualizations are constructed from a separate pool of rollout
trajectories. For each controller, we compute the final-time electric energy
and select the rollout whose value is closest to the empirical
median. This representative-case protocol avoids highlighting only unusually
stable or unstable trajectories.

All phase-space heatmaps use a shared color scale across methods and time
snapshots. Snapshots are recorded at times
\[
t \in \{0,15,30,45,60,75\}.
\]

\subsection{Implementation and Compute}
\label{app:compute-details}

The solver, training pipeline, and rollout evaluations were implemented in
Python 3.10.13 using PyTorch 2.6.0 with CUDA 12.2 support. All experiments
were executed on a university GPU cluster using single-GPU Slurm jobs.

The reported experiments used one NVIDIA L40S GPU (46\,068 MiB memory), 4 CPU
cores per task, and 64GB of RAM per job. The allocated compute nodes were
equipped with Intel Xeon Gold 6338 CPUs at 2.00\,GHz. In our Slurm setup, the
main evaluation and training jobs were run as single-GPU jobs with fixed
resource requests.

The computational cost of the overall framework is naturally divided into
three phases:
\begin{enumerate}
    \item \textbf{Expert data generation.} Offline expert datasets are produced
    by simulating the Vlasov--Poisson system under the analytical expert
    controller. On the specified hardware, generating a dataset of 400 
    expert rollout trajectories takes approximately an hour.

    \item \textbf{Simulation rollouts.} A full-horizon closed-loop rollout over
    $T=80$ constitutes the basic unit of both final evaluation and DAgger data
    aggregation. A single closed-loop rollout takes roughly 2 minutes
    to execute.

    \item \textbf{Policy training and fine-tuning.} Offline BC training and
    DAgger fine-tuning are both performed with single-GPU optimization. The BC
    training stage (100 epochs of AdamW optimization) takes approximately 
    2 hours. For DAgger, each iteration consists of a 
    rollout-aggregation phase followed by fine-tuning, taking roughly an hour per iteration.
\end{enumerate}

\section{Additional Phase-Space Rollouts}
\label{app:additional-phase-space-visualizations}

This appendix provides additional phase-space density snapshots for five rollout trajectories with independently sampled initial phases $(\phi_1,\phi_2)$. The five examples correspond to the first five random phase realizations generated using random seed 2026.

In each case, we compare the same set of controllers as in Section~\ref{sec:simulations}: uncontrolled dynamics, the analytic expert, BC-400, BC-1200, and DAgger-1. The snapshots are shown at the same time points as in Figure~\ref{fig:heatmap_bc_vs_dagger}. Across these examples, the uncontrolled and BC rollouts exhibit stronger late-time phase-space distortion and filamentation, while DAgger-1 remains closer to the expert evolution for a longer portion of the rollout.
\begin{figure*}[htbp]
    \centering
    \includegraphics[width=\textwidth]{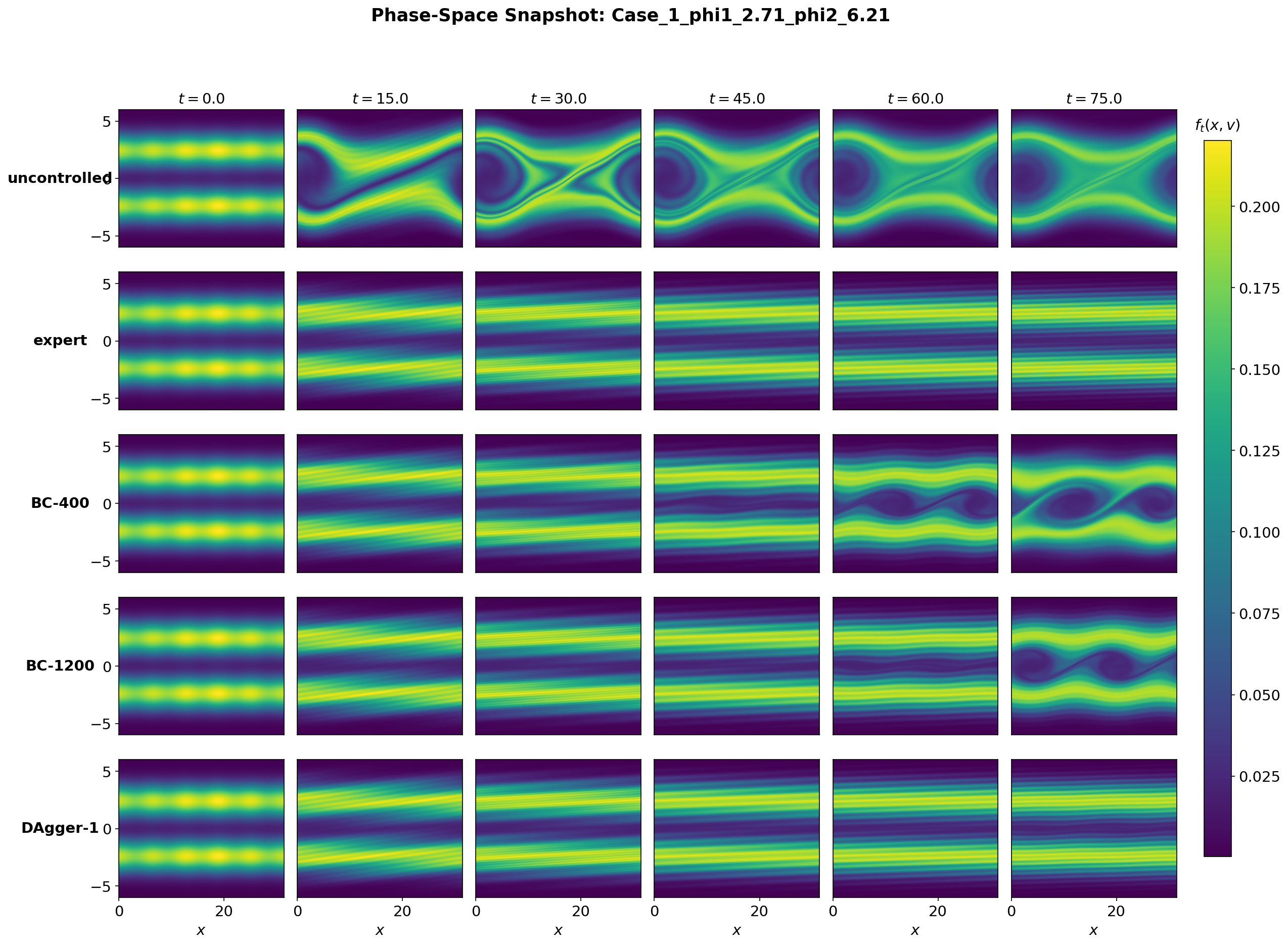}
    \caption{
    Additional rollout visualization for the first sampled initial phase pair $(\phi_1,\phi_2)$. The format matches Figure~\ref{fig:heatmap_bc_vs_dagger}.
    }
    \label{fig:app-heatmap-1}
\end{figure*}

\begin{figure*}[htbp]
    \centering
    \includegraphics[width=\textwidth]{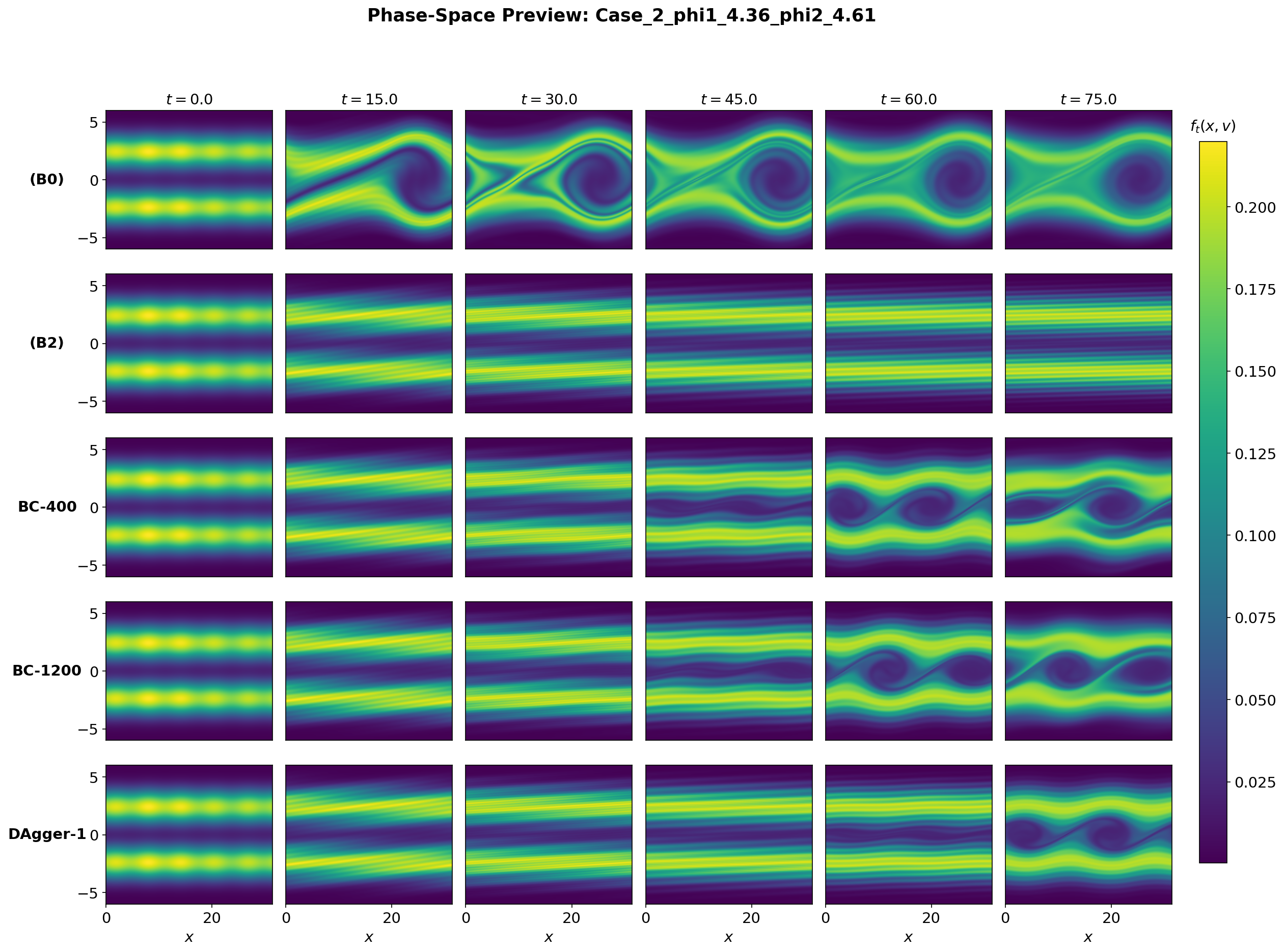}
    \caption{
    Additional rollout visualization for the second sampled initial phase pair $(\phi_1,\phi_2)$. The format matches Figure~\ref{fig:heatmap_bc_vs_dagger}.
    }
    \label{fig:app-heatmap-2}
\end{figure*}

\begin{figure*}[htbp]
    \centering
    \includegraphics[width=\textwidth]{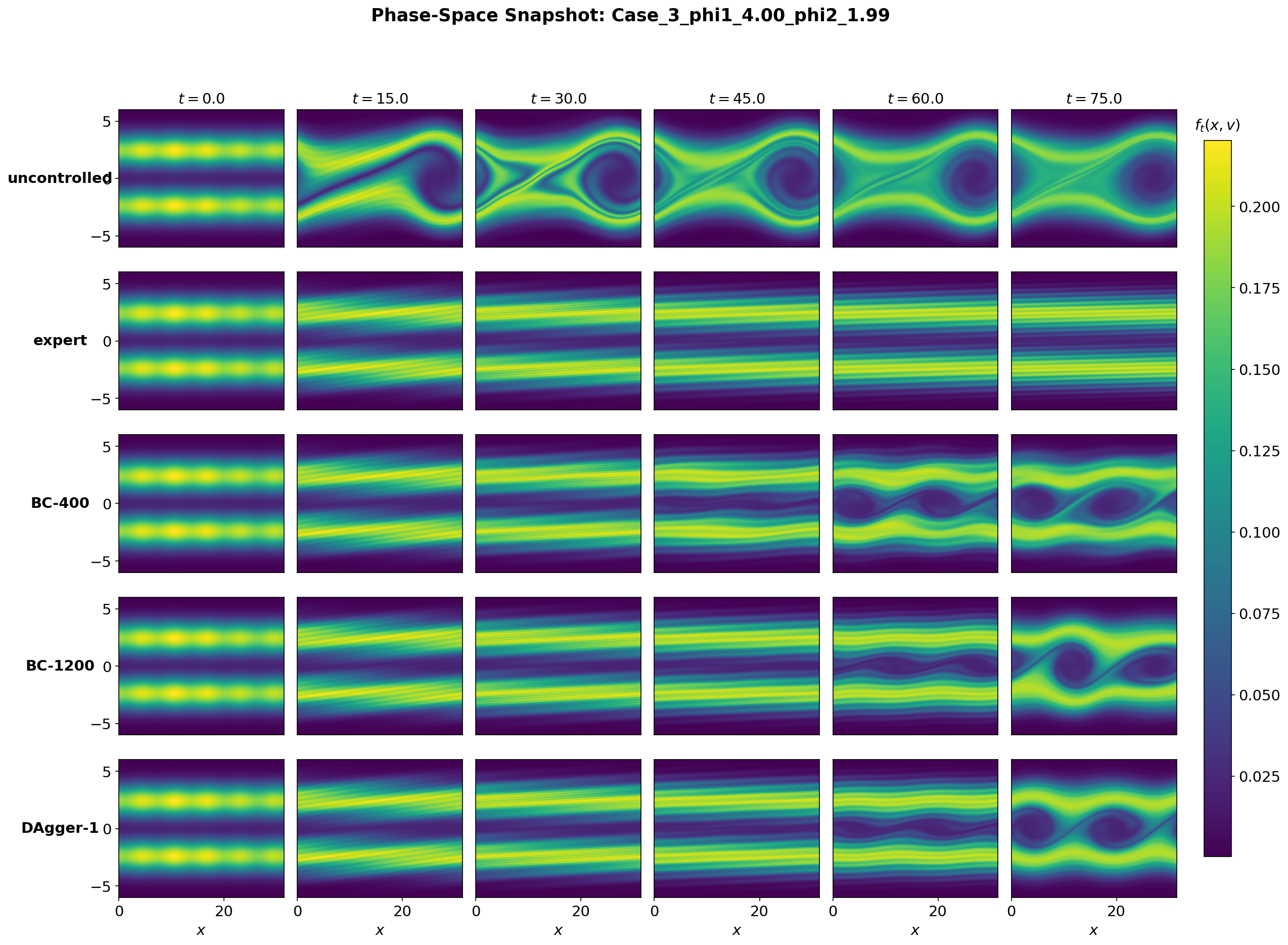}
    \caption{
    Additional rollout visualization for the third sampled initial phase pair $(\phi_1,\phi_2)$. The format matches Figure~\ref{fig:heatmap_bc_vs_dagger}.
    }
    \label{fig:app-heatmap-3}
\end{figure*}

\begin{figure*}[htbp]
    \centering
    \includegraphics[width=\textwidth]{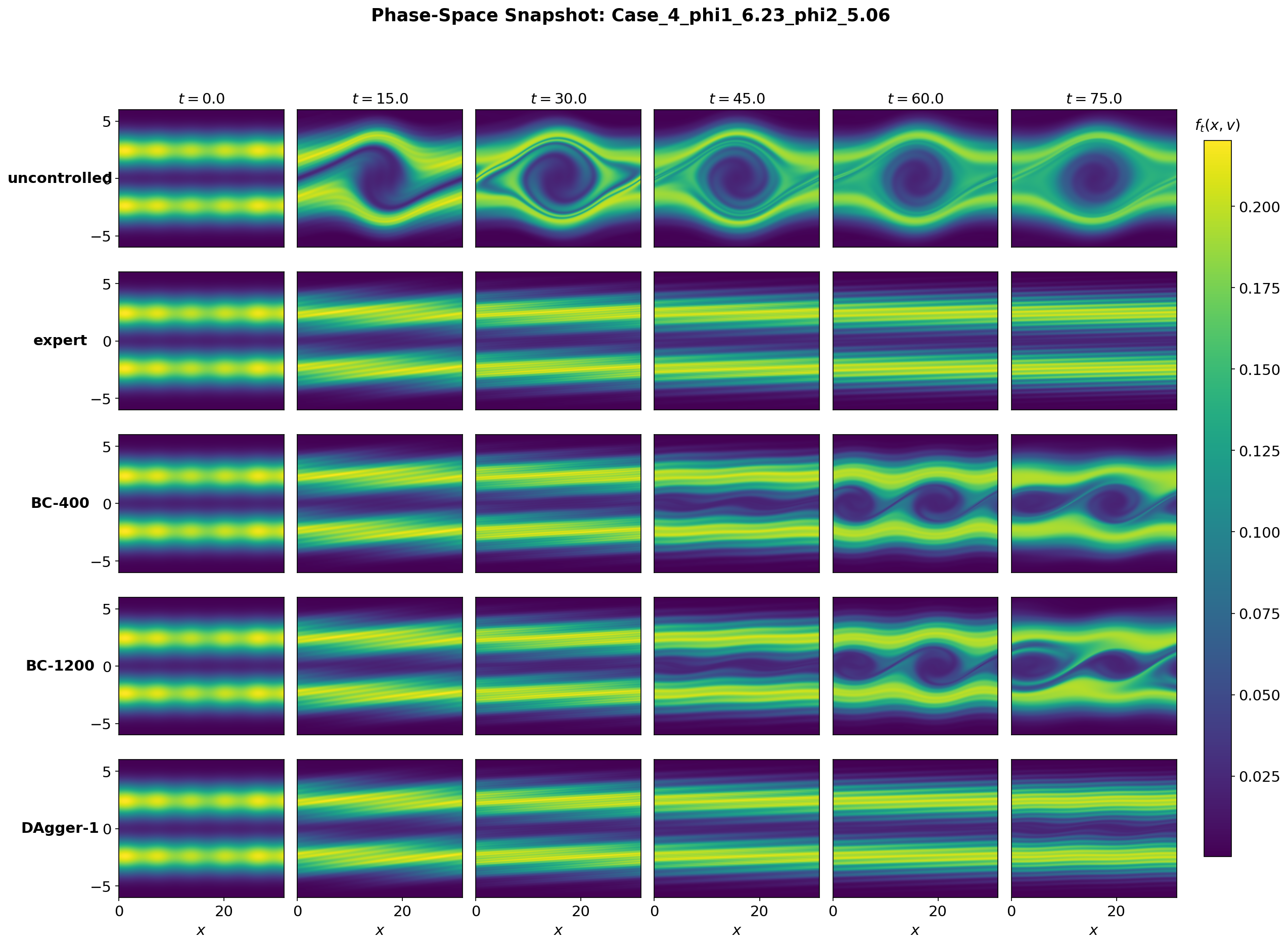}
    \caption{
    Additional rollout visualization for the fourth sampled initial phase pair $(\phi_1,\phi_2)$. The format matches Figure~\ref{fig:heatmap_bc_vs_dagger}.
    }
    \label{fig:app-heatmap-4}
\end{figure*}

\begin{figure*}[htbp]
    \centering
    \includegraphics[width=\textwidth]{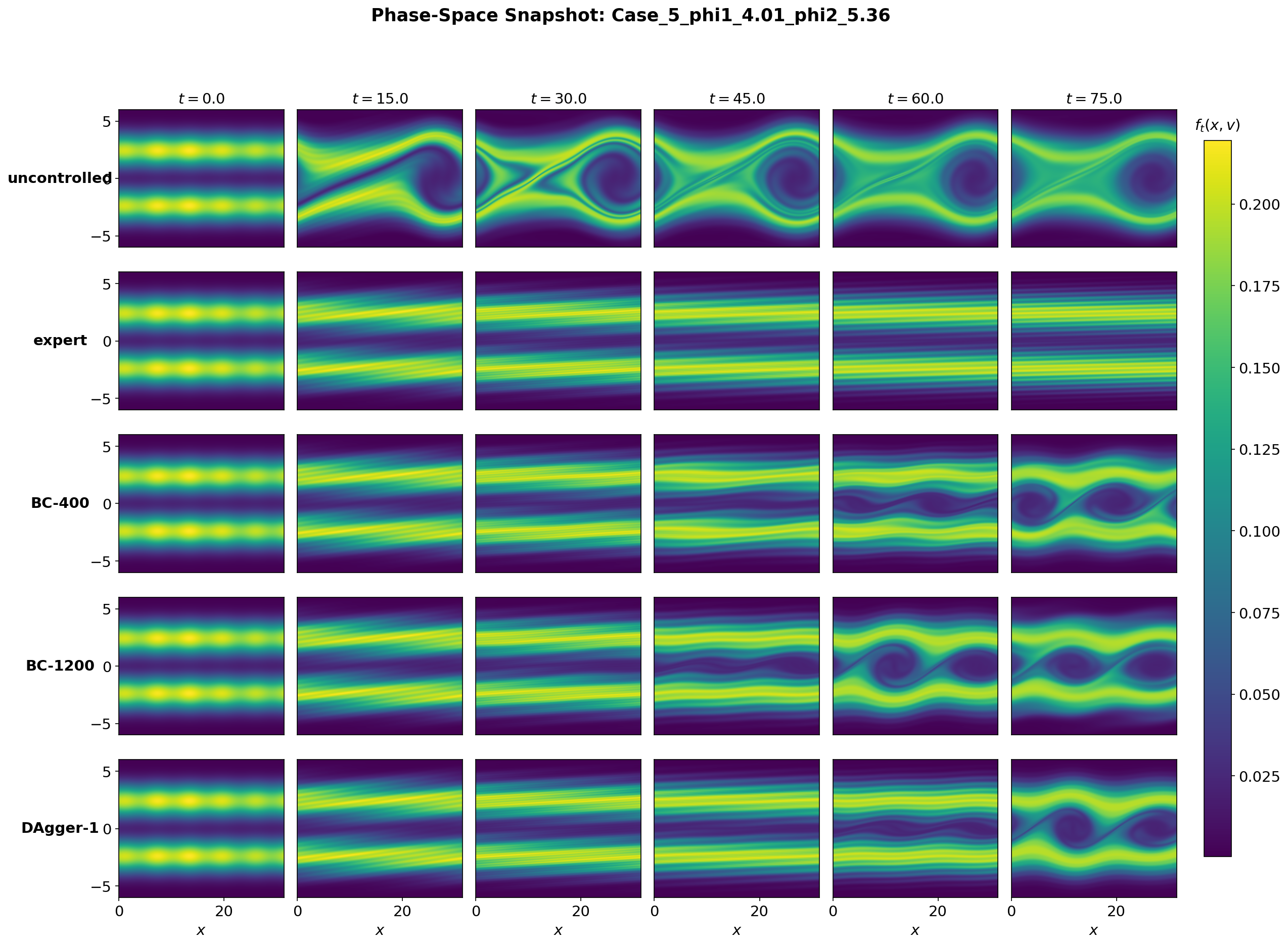}
    \caption{
    Additional rollout visualization for the fifth sampled initial phase pair $(\phi_1,\phi_2)$. The format matches Figure~\ref{fig:heatmap_bc_vs_dagger}.
    }
    \label{fig:app-heatmap-5}
\end{figure*}

\section{Additional Experimental Evaluation}
\label{app:expanded-evaluation}
In this section, we present additional experimental results to complement the evaluations provided in the main text. Specifically, Section~\ref{app:expanded-baselines} evaluates our approach against stronger baselines and investigates the impact of a second DAgger iteration, while Section~\ref{app:bump-on-tail} validates our method on the bump-on-tail instability as a complementary setting to the two-stream equilibrium.

\subsection{Expanded Paired Evaluation and Stronger Baselines}
\label{app:expanded-baselines}

We first expand the two-stream evaluation to $60$ independently sampled
initial conditions, with each initial condition paired across all
controllers. In addition to the policies in the main text, we consider two
stronger baselines designed to test whether the gain from DAgger can be
reproduced without learner-induced data aggregation.

\paragraph{Noisy-expert behavior cloning.}
This baseline broadens the offline state distribution by perturbing the
control applied during expert-data collection while retaining the clean
expert action as the label. More precisely, after the warm-up period the
rollout uses
\[
H_t^{\mathrm{rollout}}=-E_t+\xi_t,
\]
where $\xi_t$ is a zero-mean AR(1) perturbation supported on the first eight
Fourier modes, with correlation time $0.5$ and per-mode stationary standard
deviation equal to $0.002$ times the corresponding clean-target standard
deviation. The policy is trained on $400$ clean and $800$ perturbed-expert
trajectories. It therefore matches the $1{,}200$-trajectory data budget and
network architecture of DAgger-1, but its additional states are generated
without rolling out the learner.

\paragraph{Instantaneous spectral reconstruction.}
At every time step, this baseline uses only the four current density
measurements, solves the discretized Poisson equation on the four-point
sampled grid, zero-pads the resulting Fourier spectrum to the full spatial
grid, and applies $H_t=-\widehat E_t$. It has the same sensor locations and
the same prescribed control law as the learned policies, but discards the
50-step observation history. This baseline directly tests whether an
instantaneous reconstruction-then-control pipeline suffices under the sparse
observation interface.

We report two complementary late-time metrics. The first is the median
electric energy at $t=60$. The second is the mean post-transient
stabilization horizon
\[
\tau_1
=
\min\left\{
80,\,
\inf\left\{t\geq 20:\mathcal E(t)\geq 1\right\}
\right\}.
\]
Starting the threshold test at $t=20$ excludes the common initial transient,
during which even expert-controlled trajectories can temporarily cross the
order-one threshold. Larger $\tau_1$ indicates longer stabilization. The
brackets in \Cref{tab:expanded-two-stream} are $95\%$ case-level bootstrap
confidence intervals computed with $10{,}000$ resamples.

\begin{table}[htbp]
    \centering
    \small
    \setlength{\tabcolsep}{7pt}
    \begin{tabular}{lcc}
        \toprule
        Method
        & Median $\mathcal E(60)$ $\downarrow$
        & Mean $\tau_1$ $\uparrow$ \\
        \midrule
        Expert
        & $3.94\!\times\!10^{-7}$ [$3.93\!\times\!10^{-7}$, $3.95\!\times\!10^{-7}$]
        & 80.00 [80.00, 80.00] \\
        Instantaneous reconstruction
        & 2.472 [1.143, 4.013]
        & 40.24 [38.67, 42.04] \\
        BC-1200
        & 0.750 [0.648, 1.117]
        & 61.33 [59.48, 63.16] \\
        Noisy-expert BC-1200
        & 0.624 [0.405, 0.864]
        & 61.39 [58.97, 63.78] \\
        DAgger-1
        & \textbf{0.0340 [0.0151, 0.0699]}
        & \textbf{74.78 [73.31, 76.20]} \\
        \bottomrule
    \end{tabular}
    \caption{
    Expanded paired evaluation in the two-stream setting over $60$ held-out
    initial conditions. The two metrics consistently distinguish targeted
    learner-induced aggregation from increasing or generically broadening
    the offline dataset.
    }
    \label{tab:expanded-two-stream}
\end{table}

Noisy-expert BC modestly improves the terminal-energy median relative to
BC-1200, but does not extend the stabilization horizon. DAgger-1 reduces the
median terminal energy by $94.6\%$ relative to Noisy-expert BC and remains
stable substantially longer. Thus, under matched data and model budgets,
generic offline broadening does not reproduce the gain from collecting
expert labels on the states actually visited by the learner. The
instantaneous reconstruction baseline is still less stable, showing that a
single four-sensor snapshot is insufficient for reliable control and that
the temporal observation history carries essential information.

\paragraph{A second aggregation round.}
We also evaluate DAgger-2 using the same $60$ paired initial conditions and
closed-loop protocol. Its median terminal energy is
$0.0427\,[0.0183,0.0834]$, compared with
$0.0340\,[0.0151,0.0699]$ for DAgger-1. The overlapping intervals and similar
energy trajectories show no consistent improvement from the second round.
This suggests that, for the present aggregation schedule and data budget,
most of the improvement is obtained in the first round and performance has
begun to saturate; it does not imply that additional rounds cannot help under
other aggregation schedules.

\subsection{Bump-on-Tail Instability}
\label{app:bump-on-tail}

To test whether the empirical advantage of online aggregation is specific to
the symmetric two-stream equilibrium, we repeat the comparison on a distinct
bump-on-tail instability. This is a standard kinetic testbed in which a small
high-velocity population interacts with a denser background population. We
follow the one-dimensional reference configuration of
Einkemmer et al.~\cite{EINKEMMER2025113904}. Let
\[
g_{\mathrm{bg}}(v)=0.9\,\mathcal N(v;-3,1^2),
\qquad
g_{\mathrm{b}}(v)=0.1\,\mathcal N(v;4.5,0.5^2),
\]
where $\mathcal N(v;m,s^2)$ denotes the Gaussian density with mean $m$ and
variance $s^2$. The equilibrium is
$\mu_{\mathrm{bot}}=g_{\mathrm{bg}}+g_{\mathrm{b}}$, and we initialize only
the high-velocity bump with a randomized two-mode perturbation:
\[
f_0(x,v)
=
g_{\mathrm{bg}}(v)
+
g_{\mathrm{b}}(v)
\exp\!\left(
\varepsilon\left[
\cos(\beta x+\phi_1)
+
\cos(5\beta x+\phi_5)
\right]
\right),
\]
with $\beta=0.1$, $\varepsilon=0.05$, and
$\phi_1,\phi_5\stackrel{\mathrm{i.i.d.}}{\sim}\mathrm{Unif}[0,2\pi)$. We
use the periodic spatial domain $[0,2\pi/\beta]=[0,20\pi]$, velocity domain
$[-9,9]$, $n_x=n_v=1024$, time step $\Delta t=0.02$, and rollout horizon
$T=60$. Thus, this experiment changes both the equilibrium and instability
mechanism rather than merely resampling the two-stream initial condition.

The observation and action interfaces remain unchanged: four evenly spaced
density sensors, a 50-step history augmented by temporal differences, and
the first eight nonzero Fourier modes of the control field. The network
architecture is also unchanged, while all policies are trained on data from
the bump-on-tail system. BC-400 and BC-1200 use $400$ and $1{,}200$ expert
trajectories, respectively. Bump-on-tail DAgger-1 is initialized from BC-400
and uses $800$ additional aggregated trajectories; during aggregation, the
learner action is executed with probability $0.4$ and the expert action is
used otherwise. Collection is stopped after $t=20$ if the electric energy
exceeds $0.3$. Evaluation uses $60$ held-out initial conditions paired across
all controllers. As in the main experiments, control begins after a 50-step
warm-up.

\begin{table}[htbp]
    \centering
    \small
    \begin{tabular}{lcc}
        \toprule
        Instability setting
        & BC-1200
        & DAgger-1 \\
        \midrule
        Two-stream
        & 0.750 [0.648, 1.117]
        & \textbf{0.0340 [0.0151, 0.0699]} \\
        Bump-on-tail
        & 9.79 [8.40, 12.43]
        & \textbf{0.381 [0.145, 0.664]} \\
        \bottomrule
    \end{tabular}
    \caption{
    Median electric energy at $t=60$ in two distinct kinetic-instability
    settings; brackets denote $95\%$ case-level bootstrap confidence
    intervals from $10{,}000$ resamples. In the bump-on-tail experiment, the
    last stored sample is at $t=59.98$ and is reported as $t=60$.
    }
    \label{tab:bump-on-tail-comparison}
\end{table}

DAgger-1 lowers the median terminal energy relative to BC-1200 by $95.5\%$
in the two-stream setting and $96.1\%$ in the bump-on-tail setting. The
absolute energies are not directly comparable across the two equilibria, but
the relative improvement is consistent, supporting the claim that the
benefit of learner-induced aggregation is not confined to the original
two-stream configuration.

\begin{figure}[htbp]
    \centering
    \includegraphics[width=\textwidth]{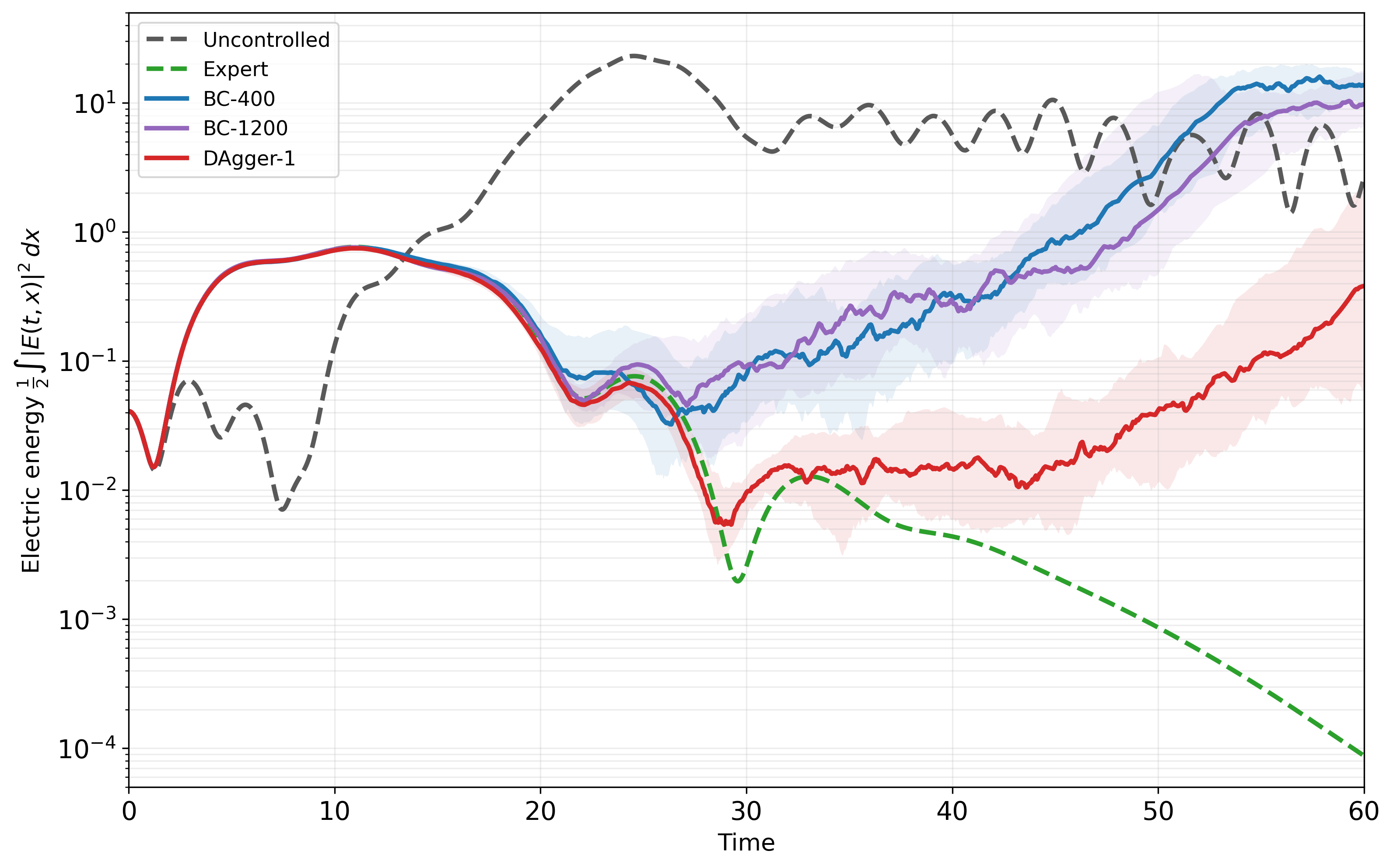}
    \caption{
    Electric-energy evolution for the bump-on-tail instability over $60$
    paired initial conditions. Curves are pointwise medians and shaded bands
    are interquartile ranges. DAgger-1 remains much closer to the expert than
    either offline BC policy over the evaluation horizon, although a gap to
    the exact expert remains at late times.
    }
    \label{fig:bump-on-tail-energy}
\end{figure}

\begin{figure}[htbp]
    \centering
    \includegraphics[width=\textwidth]{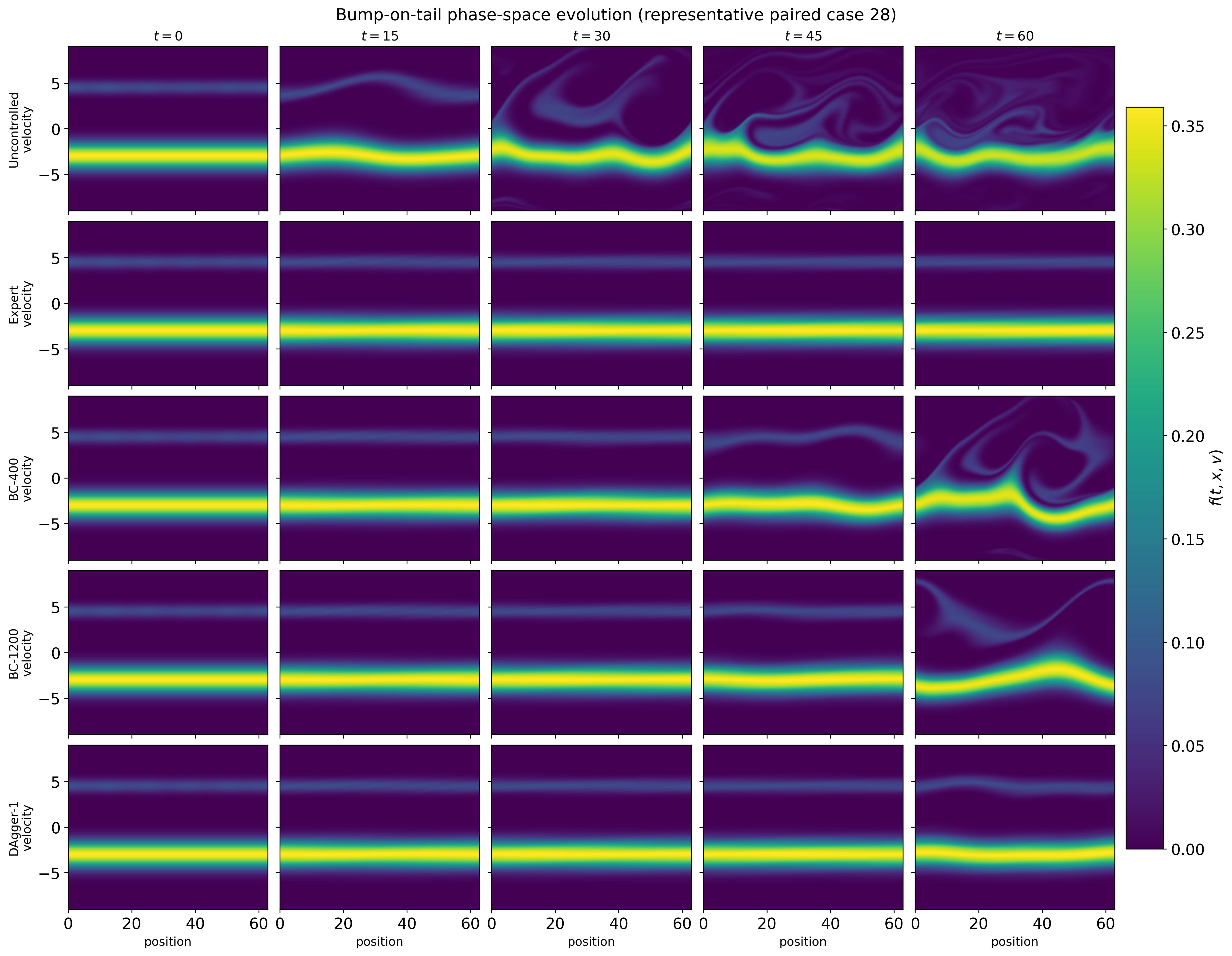}
    \caption{
    Full phase-space density $f(t,x,v)$ for a representative paired
    bump-on-tail rollout at $t\in\{0,15,30,45,60\}$. All panels share one
    color scale. The representative case is selected before inspecting the
    phase-space images by minimizing the Euclidean distance to $(0.5,0.5)$ in
    the empirical percentile-rank plane of the final BC-1200 and DAgger-1
    energies; this selects case 28. The offline policies develop pronounced
    deformation at late times, whereas DAgger-1 remains visibly closer to the
    expert evolution.
    }
    \label{fig:bump-on-tail-phase-space}
\end{figure}
\end{document}